\pdfoutput=1 
\documentclass[11pt]{article}

\usepackage[english]{babel}
\usepackage[normalem]{ulem}

\usepackage[scaled=.8]{beramono}
\usepackage[scaled=.85]{helvet}
\usepackage[T1]{fontenc}
\usepackage{fourier}
\usepackage{color,soul}

\usepackage[round]{natbib}

\usepackage[colorlinks=true,linkcolor=black,citecolor=black,filecolor=black,urlcolor=black]{hyperref}

\usepackage{color}
\newcommand{\be}{\begin{enumerate}}
\newcommand{\ee}{\end{enumerate}}
\newcommand{\german}[1]{{\color{red}#1$^{de}$}}

\newcommand{\spanish}[1]{{\color{red}#1$^{es}$}}
\newcommand{\hebrew}[1]{{\color{red}#1$^{he}$}}

\newcommand{\highlight}[1]{{#1}}

\newcommand{\orig}[1]{}

\newcommand{\dcom}[1]{}

\newcommand{\dout}[1]{}

\newcommand{\oa}[1]{{\color{green}{OA: #1}}}
\newcommand{\nss}[1]{{\color{magenta}{NSS: #1}}}

\newcommand{\rem}[1]{{(\it #1})}
\newcommand{\imp}[0]{\rem{IMP}}

\usepackage[top=0.5in, bottom=1.5in, left=0.5in, right=0.5in]{geometry}

\usepackage{authblk}
\title{UCCA's Foundational Layer:  Annotation Guidelines v2.1 
}

\author[1]{Omri Abend}
\author[2]{Nathan Schneider}
\author[1]{Dotan Dvir}
\author[2]{Jakob Prange}
\author[1]{Ari Rappoport}
\affil[1]{Hebrew University of Jerusalem}
\affil[2]{Georgetown University}

\date{December 31, 2020}

\begin{document}
\maketitle

\begin{abstract}
\noindent
    This is the annotation manual for Universal Conceptual Cognitive Annotation \citep[UCCA;][]{ucca}, specifically the Foundational Layer.
    UCCA is a graph-based semantic annotation scheme based on typological linguistic principles.
    It has been applied to several languages; for ease of exposition these guidelines give examples mainly in English.
    New annotators may wish to start with the tutorial on the UCCA framework \citep{tutorial}. 
    Further resources are available at the project homepage: \url{https://universalconceptualcognitiveannotation.github.io}
\end{abstract}

\tableofcontents

\newpage

\section{General Principles for Annotation}

\begin{enumerate}

\item
  A UCCA annotation task consists of the annotation of multiple sentences, usually a 
  paragraph or several paragraphs long.
  When you receive a task, take a few minutes to read the entire text, in order to 
  understand the context.
  
 \item
  UCCA divides the text into units (stretches of text; not necessarily contiguous), each referring to a relation,
  a participant in a relation or a relation along with its participants. The types of relations we annotate
  are listed below.






\item
  The units must cover all the tokens, except punctuation tokens which are not annotated.
  
\item
  Units may contain other sub-units, giving rise to a hierarchical structure.
  
\item
  Each unit is assigned a category, reflecting its role in a super-ordinate relation in which it participates. The category does not necessarily reflect the meaning of the unit taken in itself. For instance, all the units in {\bf boldface} have the same category, as they all describe ``horse'' in finer detail:

\begin{itemize}
\item
  ``A {\bf beautiful} horse''
\item
  ``A {\bf police} horse''
\item
  ``A horse {\bf with no name}''
\item
  ``The horse {\bf that won the race}''
\item
  ``A {\bf winning} horse''
\end{itemize}

\item
  UCCA does not annotate ambiguity. 
  When reading ambiguous text, decide on the most likely interpretation in your opinion and use it for annotating
  the entire passage.

\end{enumerate}

\section{A Bird's Eye View of the Categories}

Units may be analyzed according to {\bf one} of the following models:

\subsection*{Model \#1: Scenes}\label{model1}

\begin{enumerate}
  \item
    A Scene is some description of an action, movement or state (including abstract actions or states).
    It generally has a time when it happened, a location, and a ground (the circumstances in which the description was
    uttered or written).
    In concrete cases, a Scene can be imagined as a single mental image or a short script.
    \begin{itemize}
    \item
      ``Woody walked in the park'' (1 Scene)
    \item
      ``I got home and took a shower'' (2 Scenes)
    \end{itemize}

  \item
    A Scene has a main relation (exactly one), which determines the type of the Scene and what happened in it. This main relation can be either a {\sc State} (S) -- if the Scene is constant in time; or a {\sc Process} (P) -- an action, movement or some other relation that evolves in time.
     
  \item
    Each Scene is considered a unit, and is therefore, like all units, also assigned a category as a whole. The category of the Scene unit reflects the role of that unit in the super-ordinate relation it participates in (see below).

  \item
Scenes may contain any number of {\sc Participants} (A).  These are the principal participants in the main relation of the Scene (including locations). Participants may refer either to physical or abstract entities.

  \begin{itemize}
  \item
    ``John$_A$ boiled [an egg]$_A$''
  \item
    ``Programming$_A$ is widely taught nowadays''
  \end{itemize}

  \item
    In static Scenes, the main relation is annotated as a {\sc State}. The State unit should not include its auxiliary verbs if present.

    \begin{itemize}
    \item
      ``John$_A$ is tall$_S$''
    \item
      ``[The apple tree]$_A$ is in$_S$ [the garden]$_A$''
    \item
      ``[A truck]$_A$ weighs$_S$ tons$_A$''
    \item
      ``[This apple]$_A$ does weigh$_S$ [200 g]$_A$''
    \item
      ``[Big$_S$ \rem{dogs}$_A$]$_E$ dogs$_C$'' 
    \end{itemize}
    
    
  \item
    In dynamic Scenes, the main relation is marked as a {\sc Process} (P). The Process unit should not include its auxiliary verbs if present.
    
    \begin{itemize}
    \item
      ``John$_A$ kicked$_P$ [the ball]$_A$''
    \item
      ``John$_A$ has been kicking$_P$ [the ball]$_A$ since breakfast''
    \end{itemize}
    
  \item
    {\sc Adverbials} (D) are relations that do not introduce another Scene, but semantically modify the Scene or its {\sc Process} or {\sc State}. Common cases of Ds are modal relations (like ``can'', ``may'' or ``want''), manner relations (like ``quickly'' or ``patiently'') and relations that specify a sub-event (like ``begin'' or ``finish'').
    
    \begin{itemize}
    \item
      ``John$_A$ began$_D$ swimming$_P$''
    \item
      ``John$_A$ may$_D$ come$_P$ [to the party]$_A$''
    \item
      ``[His workers]$_A$ treat$_P$ him$_A$ [with disrespect]$_D$''
    \item
      ``John$_A$ cleverly$_D$ answered$_P$ [the manager's question]$_A$''
    \end{itemize}
    
    See Section~\ref{app:AD-distinction} for how to distinguish Ds and As in edge cases.

  \item
    Units whose primary purpose is to specify the time in which the Scene occurred should be marked as {\sc Time} (T). However, if time is described by introducing another Participant or another Scene, it should receive a Scene or Participant category instead. Frequency and duration are also considered {\sc Time}.
    
    \begin{itemize}
    \item ``John$_A$ may$_D$ come$_P$ later$_T$''
    \item ``John$_A$ may$_D$ come$_P$ [at around eight]$_T$''
	\item ``I$_A$ get$_F$ treated$_P$ regularly$_T$''
	\item ``[John$_A$ [showed up]$_P$]$_H$ during$_L$ [[the$_F$ filming$_C$]$_P$]$_H$'' (two Scenes! see below)
    \end{itemize}

\end{enumerate}

\subsection*{Model \#2: Non-Scene Units}\label{model2}


In some cases a unit can be internally analyzed, but its elements do not evoke a scene. We distinguish between several types of non-Scene relations:

\begin{enumerate}
\item
An element will be marked as {\sc Center} (C) when it is necessary for the conceptualization of the non-Scene unit. It is an element on which other relations may elaborate or connect. Each non-Scene unit should include at least one Center.

\begin{itemize}
\item 
Queen Elizabeth$_C$
\end{itemize}

\item
{\sc Elaborators} (E) add some information on the main element (Center).  These either include inherent attributes (attributes that cannot change because they define the element) or one of several types of relations specified below.

  \begin{itemize}
    \item
      ``Queen$_C$ [of$_R$ England$_C$]$_E$'' (describes a type of a queen; the fact that     	she is the queen of England is inherent to her being a queen)
        
	\item
	  ``His$_E$ hand$_C$''
	\item
      ``Chocolate$_E$ cookies$_C$'', ``Metal$_E$ belt$_C$'' (the substance something is made of)
  \end{itemize}
  
  Another type of relations that are considered Es are class descriptor: units 
  comprised of a sub-unit that specifies the name of the entity in question, and 
  another unit specifies which category it belongs to. In these cases, the specific 
  unit is marked with $C$ and the class descriptor is marked as $E$.
  
		\begin{itemize}
		\item
			``[the name]$_E$ [John]$_C$''
		\item
			``[the state]$_E$ [of$_R$ Washington$_C$]$_C$''
		\item
			``[the year]$_E$ [1966]$_C$''
        \end{itemize}

Note that if it is hard to say which of the sub-units adds information to which, both units should be marked as Cs. That is, if there is no one unit that determines the type of entity, all units that determine its type should be marked as Cs.
  A frequent example of that is part-whole relations: Units comprised of a sub-unit that specifies the whole, and one that specifies a sub-part of it. 

  \begin{itemize} 
  \item
    ``bottom$_C$ [of$_R$ the$_F$ sea$_C$]$_C$''
  \item
  ``sea$_C$ bottom$_C$''
    \item
    ``tip$_C$ [of$_R$ the$_F$ iceberg$_C$]$_C$''
   \item
	``the$_F$ [iceberg$_C$ 's$_R$]$_C$ tip$_C$''
  \german{\item
    ``See-$_C$ boden$_C$''
  } 
  \end{itemize}

\item
 {\sc Connectors} (N) relate two or more entities (annotated as Centers) in a way that 
 highlights the fact that they have a similar type or role.
 They are usually conjuncts such as the English ``and'', ``or'', ``instead of'' or ``except'' or the German ``sowie'', ``ebenso'' and ``genauso wie''.

  \begin{itemize}
  \item
    ``[John$_C$ and$_N$ Mary$_C$]$_A$ went$_P$ [to school]$_A$ together$_D$''
  \item
    ``I$_A$ 'll$_F$ have$_P$ [coffee$_C$ and$_N$ cookies$_C$]$_A$'' 
  \end{itemize}

\item 
We use {\sc Quantifiers (Q)} to mark expressions that:
 
\begin{itemize}
\item
Describe the quantity or magnitude of an entity:

\begin{itemize}
\item
``three$_Q$ apples$_C$''
\item
``several$_Q$ apples$_C$''
\item
``I$_A$ bought$_P$ [[three$_Q$ kilos$_C$]$_Q$ [of$_R$ apples$_C$]$_C$]$_A$'' 
\end{itemize}

\item 
Any expression that defines that an entity is a group or a set (e.g., ``group of ...'', ``hundreds of ...'').

\begin{itemize}
\item
``[a$_F$ group$_C$]$_Q$ [of$_R$ people$_C$]$_C$''
\item 
``[a$_F$ swarm$_C$]$_Q$ [of$_R$ bees$_C$]$_C$''
\item  
``[a$_F$ variety$_C$]$_Q$ [of$_R$ colors$_C$]$_C$''
\end{itemize}
\end{itemize}

\end {enumerate}
\subsection*{Model \#3: Inter-Scene relations}

\begin{enumerate}
  
\item
  Linkage is the term for inter-Scene relations in UCCA. There are four major types of 
  relations in
  which Scenes may participate, and therefore four types of categories Scene units may 
  receive.
  The next four items describe these types. 
  
\item
  \textbf{Elaborator Scenes:} an E-Scene adds information to a previously established 
  unit. Usually answers
  a ``which X'' or ``what kind of X'' question. Es should place the C they are 
  elaborating as a {\it remote} A (see below).
  A way to check where a Scene is an E-Scene is to ask whether the Scene along with the C 
  it relates to are
  of the same type as the C itself.

  \begin{itemize}
  \item
    ``[The$_F$ dog$_C$ [that ate my homework \rem{dog}$_A$]$_{E}$ ]$_A$ is brown'' 
    (``dog'' is a remote A in ``that ate my homework'')
  \item
    ``The$_F$ person$_C$ [whom$_R$ I$_A$ gave$_P$ [the present]$_A$ [to$_R$ 	
        \rem{person}$_C$]$_A$ ]$_{E}$''
  \item
    ``Brad played [an$_F$ American$_C$ [taken to the Adriatic 	
    	\rem{American}$_A$]$_{E}$]$_A$''
  \end{itemize}

\item
  \textbf{Participant Scenes:} an A-Scene is a participant in a larger Scene. It does not add information to some specific participant in it, and if you remove it, it doesn't retain the same type. Usually answers a ``what'' question about the Scene.
  
  \begin{itemize}
  \item
    ``[Talking to strangers]$_A$ is$_F$ ill-advised$_S$'' (answers ``what is ill-advised?'')
  \item
    ``John$_A$ said$_P$ [he's hungry]$_A$'' (answers ``what did John say?'')
  \item
    ``[[John$_C$ 's$_R$]$_A$ accurate$_D$ kick$_P$]$_A$ saved$_P$ [the game]$_A$'' (answers ``what saved the game?'')
  \end{itemize}
  
\item \textbf{Center Scenes:} a C-scene is when a unit is first marked as the Center unit of a larger scene and then internally annotated as scene. These should be used sparingly and as a last resort. In most cases we can avoid C-scenes by marking the scene-evoking Center unit directly as a P/S:

\begin{itemize}
\item 
``[the$_F$ experienced$_D$ judge$_{P+A}$]$_A$ knew how to deal with the difficult case.'' (since `judge' evokes a Scene and `experienced' modifies the process of judging, we can mark the whole unit directly as a scene instead of first marking `judge' as C)
\item
``[the$_F$ taxi$_A$ driver$_{P+A}$]$_A$ went home'' (also here the unit can be annotated directly as a scene)
\item
``a$_F$ lung$_A$ specialist$_{P+A}$''
\end{itemize}


When are C-scenes necessary?
In cases where the Center unit in a phrase is scene-evoking but the phrase itself cannot be annotated directly as a scene:

\begin{itemize}
\item 
``[The$_F$ [tall$_S$ \rem{judge}$_A$]$_E$ [judge$_{P+A}$]$_C$]$_A$ reached for the book on the top shelf'' (since `tall' doesn't modify the process of judging, we can't annotate the whole unit (``the tall judge'') directly as a scene and so will have to first mark judge as C) 
\item
``the$_F$ [taxi$_A$ driver$_{P+A}$]$_C$ [who$_R$ swims$_P$ \rem{driver}$_A$]$_E$'' (also here a C-scene is necessary, since the unit ``who swims'' doesn't modify the process of driving) 
\item
``[[the$_F$ [tall$_S$ \rem{teacher}$_A$]$_E$ [English$_A$ teacher$_{P+A}$]$_C$]$_A$ arrived$_P$]$_H$''
\end{itemize}

\item
  {\bf Parallel Scenes:} any other Scene receives the category Parallel Scene (H). 
  Sometimes there is an accompanying relation word and sometimes not. If so, it is a 
  Linker (L). Note that there are no Adverbial (D) Scenes or Time (T) Scenes.
  Except for Ground (see below), if a Scene is not an A (Participant), a C (Center) or an E (Elaborator), it's an H.

 \begin{itemize}
  \item
    ``[John managed to amuse himself]$_H$ while$_L$ [waiting in line 
    	\rem{John}$_A$]$_H$''
	\item
    ``[The minute]$_L$ [I got home]$_H$ [I noticed the new painting]$_H$''
  \item
    ``If$_L$ [you build it]$_H$ [they will come]$_H$'' 
  \item
    ``[I've done some research]$_H$, [asked a couple of questions \rem{I}$_A$]$_H$ and$_L$ [started thinking \rem{I}$_A$]$_H$''
  \item
    ``[You're only saying this]$_H$ because$_L$ [John$_A$ told$_P$ you$_A$ to$_F$ \rem{You're only saying this}$_A$ ]$_A$ ]$_H$'' 
    
  \german{\item
    ``Nach$_L$ [einer Rolle in einem Thriller]$_H$ [spielte sie in einem Actionfilm mit]$_H$.''}
  \end{itemize}



  Specific cases of Parallel Scenes include (examples of relevant Linkers in brackets): purposive (``in order to'' or ``to''\german{, `um + zu-Infinitiv''}), logical (``if ... then ...''), temporal (``when X, Y'', ``before X, Y''), coordination (``and'', ``but''), and contrastive linkages (``however'', ``still''\german{, ``jedoch''}).

Scenes that are not related to any other units and are therefore in the top level of organization in the text are also Hs (Parallel Scenes). 

Linkers do not necessarily appear between the Scenes they are linking (see example \#2 above).

\item
  A unit is marked as {\sc Ground} (G), if its primary purpose is to relate some unit to its speech event; either the speaker, the hearer or the general context in which the text was uttered/written/conceived.\footnote{The speech event is called Ground following R. Langacker.}

  Gs are similar to Ls, except that they don't relate the Scene to some other Scene in the text, but rather to the speech act
  of the text (the speaker, the hearer or their opinions). By convention,
  Ground units should be positioned within the Scene they relate to.

  \begin{itemize}
  \item
    ``[Surprisingly$_G$ , [our flight]$_A$ arrived$_P$ [on time]$_T$]$_H$''
  \item
    ``[[In my opinion]$_G$, John$_A$ is$_F$ coming$_P$ home$_A$ ]$_H$'' 
  \end{itemize}
  
{\bf Note:} a complete Scene that refers to the ground (with As and Ds etc.) should be annotated as a Scene and not as a G. That is, if a unit alludes to the speech event, but is missing almost all its elements save for one word or expression, it should be a G. If the speech event is mentioned more elaborately, it should be annotated as a Scene.

  \begin{itemize}
  \item
    ``[I$_A$ was$_F$ surprised$_S$ ]$_H$ when$_L$ [[our flight]$_A$ arrived$_P$ [on time]$_T$]$_H$''
  \item
    But: ``[Surprisingly$_G$, [our flight]$_A$ arrived$_P$ [on time]$_T$]$_H$''
  \item
    ``I$_A$ told$_P$ you$_A$ already$_D$ [that John can't make it]$_A$''
  \end{itemize}
  
\end{enumerate}

\subsection*{Categories that Appear in All Models}

There are two types of categories that may appear anywhere in the text:  Relators (R), Functions (F).


\begin{enumerate}
\item 
Relators (R) are relations that relate two or more entities within Scene units as well as non-Scene units. Rs in English are usually prepositions (see Section~\ref{sec:relators} below for a more elaborate discussion).

We place Rs inside the unit they pertain to:

\begin{itemize}
\item
``John said [that$_R$ he$_A$ 's$_F$ going$_P$ home$_A$]$_A$''
\item
``John$_A$ was$_F$ sick$_S$ [on$_R$ Monday$_C$]$_T$'' 
\item
``John$_A$ left$_P$ [in$_R$ a$_F$ hurry$_C$]$_D$''
\item 
``John$_A$ met$_P$ [with$_R$ [Mary$_C$ and$_N$ Jane$_C$]$_C$]$_A$''
\item
``[The end]$_D$ [of$_R$ the war]$_P$''
\item
``Queen$_C$ [of$_R$ England$_C$]$_E$''
\item
``Plenty$_Q$ [of$_R$ money$_C$]$_C$''
\end{itemize}


 

  
	When will we {\bf not} use Relators?: 
    \begin{itemize}
    \item 
    To link between Parallel Scenes (for that see Linkers).
    \item
    To connect between Centers that have the same parent unit and carry 
    a similar type or role (for that see Connectors). 
  \end{itemize}





    
    


\item
  Functions (F) are units that do not introduce a new participant or relation.
  They can only be interpreted as part of a larger construction in which they are 
  situated, or convey some aspect of meaning which is not covered by the foundational 
  layer (e.g., tense or focus). Usually in these cases, they cannot be substituted 
  with any other word.

  \begin{itemize}
  \item
    ``I$_A$ want$_D$ to$_F$ run$_P$ [a$_F$ marathon$_C$]$_A$''
  \item
    ``I$_A$ am$_F$ going$_P$ [to$_R$ the$_F$ supermarket$_C$]$_A$''
  \item
    ``It$_F$ is likely$_D$ that$_F$ John$_A$ will$_D$ come$_P$''
  \item
    ``Let$_F$ me$_A$ introduce$_P$ John$_A$''
  \end{itemize}
  
  

\end{enumerate}

\subsection*{Remote and Implicit Units}

\begin{itemize}
\item
There are instances where a sub-unit in a given unit is not explicitly mentioned. We can indicate the missing sub-unit in two ways:
\be
\item 
Add a reference of the missing unit from another place in the text, as a Remote unit. By convention, the remote unit should be selected to be the minimal unit that refers to the target entity (for instance, ``table'' and not ``the red table''). 

Usually the minimal unit is a single word or unanalyzable unit, though occasionally a larger remote is needed to avoid a unit whose children consist only of remote and F units, as in:\\
``[You're only saying this]$_H$ because$_L$ [John$_A$ told$_P$ you$_A$ to$_F$ \rem{You're only saying this}$_A$ ]$_A$ ]$_H$'' 
\item When it does not appear explicitly in any place in the text, add an Implicit unit to stand for the missing sub-unit.  \ee

Note that remote and implicit units should be assigned relevant categories like any other unit.

\item
Since it's often difficult to determine when units that do not appear explicitly in the text (i.e., remotes/implicits) do in fact play a role, we add remotes/implicits only in cases where the remotes/implicits are within a certain grammatical structure that licenses their omission.\footnote{This is called ``constructional null instantiation'' in  frame semantics \citep{ruppenhofer-16}.} For instance, imperatives in English do not require mentioning the person being ordered, but that person is saliently present in the Scene. 

Other cases of omission are not related to the grammatical structure, but to an inference based on the words themselves. For instance, ``he came late'' implies that there's a place he came to, but we know that based on the semantics of ``came'', rather than by the grammar. In this case we do not add a remote/implicit.


Another case where remotes/implicits are needed is if the unit lacks a Center or a Process/State.

Note that \citet{cui-20} offered a proposal for richer treatment of implicits, but this would belong in a new extension layer and is not part of the Foundational Layer.

Below is a list of prominent examples of cases that require adding a remote/implicit
(implicits/remotes in bold):

\begin{enumerate}
\item 
Imperative: 
``[Come$_P$ here$_A$, \imp$_A$]$_H$'' (who should come here?)
\\ Note that if the party addressed appears explicitly in the text we should annotate it as a Remote instead of Implicit (see Section~\ref{imperatives}).
\item
Passive:
``[[the computer]$_A$ was$_F$ stolen$_P$ \imp$_A$]$_H$'' (by whom was the computer stolen?)
\item
Coordination between clauses that share the same subject:
``[John came home]$_H$ and$_L$ [ate \rem{\textbf{John}}$_A$]$_H$''
\item
Inferred subject constructions: 
\begin{itemize}
\item
``[Born$_P$ [to a conservative household]$_A$, \rem{\textbf{Pitt}}$_A$]$_H$, [Pitt$_A$ was$_F$ sent$_P$ [to a Catholic school]$_A$]$_H$'' 
\item
``To$_L$ [win$_P$ \rem{\bf you}$_A$]$_H$, [you$_A$ [have to UNA]$_D$ find$_P$ [the$_F$ key$_C$]$_A$]$_H$''
\item
Gerunds: ``[waiting$_P$ \imp$_A$ [for$_R$ the$_F$ doctor$_{P+A}$]$_A$]$_A$ can$_D$ be$_F$ boring$_S$''
\end{itemize}

\item
Gapping:
``[John$_A$ bought$_P$ eggs$_A$]$_H$, [Mary$_A$ gum$_A$, \rem{\textbf{bought}}$_P$]$_H$''
\item
Ellipsis:
``[You bought three horses]$_H$, [I bought [one$_Q$, \rem{\textbf{horses}}$_C$]$_A$]$_H$'' (``horse'' is inferred)
\item
Relative clauses:
``[[The$_F$ dog$_C$ [I$_A$ saw$_P$ [last night]$_T$, \rem {\textbf{dog}}$_A$]$_E$]$_A$ barked$_P$]$_H$''
\item
Infinitive clauses: 
``[I$_A$ told$_P$ him$_A$ [to$_F$ go$_P$, \rem{\textbf{him}}$_A$]$_A$]$_H$'' (``him'' is object of one and subject of another)
\item
Subject omission (in languages that allow it):\\
\spanish{``No$_D$ hablo$_P$ Ingles$_A$ \imp$_A$''} (the speaker is implicit)\\
gloss: not speak$_{1sg,pres.}$ English\\
translation: I don't speak English

\end{enumerate}

Note that we do not add Fs as remotes and we usually do not add Rs as remotes unless in rare cases of ellipsis. 

Examples:
\begin{itemize}
\item
``John will go [to$_R$ [Paris and London]$_C$]$_A$'' (single A with single R, remote is not needed)
\item
``John will go to Paris and [so$_D$ will$_F$ \rem{go}$_P$ \rem{to$_R$ Paris$_C$}$_A$ Mary]$_H$'' (entire A is added as remote)
\item
``John will go to Paris and Mary, London'':
``...and$_L$ [Mary$_A$, \rem{go}$_P$ [\rem{to}$_R$ London$_C$]$_A$]$_H$'' (R is added as remote)
\end{itemize}


\end{itemize}

\section{Technical Notes and Conventions}\label{sec:technical_notes}

\begin{enumerate}
\item
With any problem or question, contact the administrator of the project. 
If there is uncertainty, mark your guess and add ``uncertain''.
\item
When annotating a remote unit, select the minimal possible relevant unit, and not its ancestors.
\item
Top-level annotation (i.e., of units directly below the passage level) should be annotated, wherever possible, according to the Scene model. The only exceptions are cases that do not describe a Scene in any way (such as section titles).
\item
Prefer Ls over Ds, where possible.
\item
Prefer Ls over Gs where possible.
\item
Prefer Ls over Ts where possible.
\item
R and F are residual categories, and so should not be marked if other categories are applicable.
\item
Prefer annotating A-Scenes and E-Scenes over Parallel Scenes where possible.
\item
Prefer separating participants from their P/S where possible.
\item
Prefer Ds over a longer P/S with an E inside it. More generally, try to avoid complex or long P/S.
\item
Use Implicit units sparingly and prefer Remote units where possible.
\item
Do not create units only to be used later as a Remote unit. Use existing units instead.
\item
Since morphology in English is very impoverished, we take a pragmatic approach and in our primary layer do not annotate parts of words, but only complete words. 
\item
Function units (Fs) do not refer to a participant or relation and, since the UCCA annotation reflects participation in relations, it is often not clear in what level of the hierarchy an F unit should be placed in. When this occurs, include the F in the deepest unit that stands to reason.
\item
Single words are often Scenes as well. This will usually happen where none of the participants is explicitly mentioned.

\begin{itemize}
\item
  ``I$_A$ remember$_P$ [the$_F$ negotiations$_P$]$_A$''
\item
  ``The$_F$ [available$_S$ \rem{options}$_A$]$_E$ options$_C$''  
\item
 ``[Crying$_P$ \rem{you}$_A$ ]$_A$ makes$_D$ you$_A$ stronger$_P$''
 \item
 ``[I$_A$ went$_P$ [to$_R$ the$_F$ store$_C$]$_A$]$_H$ for$_L$ [eggs$_A$ \rem{I}$_A$ \imp$_P$]$_H$'' (``for'' is a purposive linker. The implicit P in in the second Scene is for the buying action)
\end{itemize}

\end{enumerate}

\section{Classification of Prepositions}\label{sec:relators}

Prepositions are in frequent use in English. They include words such as ``in'', ``on'', ``after'', ``with'' and ``under'' \german{ or ``nach'', ``in'' and ``auf'' in German}.
Some prepositions are multi-worded, in which case they are internally annotated as unanalyzable. Examples include ``thanks to'' and ``on top of''.

\begin{enumerate}
 
\item
{\bf Prepositions as Relators:} Both in Scene units and in non-Scene units, Relators should be included inside the unit they pertain to.
\begin {enumerate} 
\item Scene unit examples:
\begin {itemize}
	\item
	``John$_A$ put$_P$ [the$_F$ hat$_C$]$_A$ [on$_R$ the$_F$ shelf$_C$]$_A$''
	\item
	``John$_A$ relied$_P$ [on$_R$ his$_A$ father$_{S+A}$]$_A$''
	\item
	``John$_A$ referred$_P$ [to$_R$ Mary$_C$]$_A$ [in$_R$ his$_A$ dissertation$_P$]$_A$''
    \item
    ``he$_A$ left$_P$ [in$_R$ a$_F$ hurry$_C$]$_D$''
    \item
    ``[His book]$_A$ was$_F$ published$_P$ [in$_R$ 2014$_C$]$_T$''
    \item
    ``John$_A$ will$_F$ visit$_P$ Mary$_A$ [on$_R$ Sunday$_C$]$_T$''
\end {itemize}

\item Non-Scene unit examples: (where a preposition and its object serves as an Elaborator of a non-Scene unit, the preposition is invariably R)
   \begin{itemize}
	\item
	``President$_C$ [of$_R$ the$_F$ USA$_C$]$_E$''
	\item
	``bottom$_C$ [of$_R$ the$_F$ sea$_C$]$_C$''
	\item
	``[a period]$_C$ [of$_R$ time$_C$]$_C$''
	\item	
	``[a group]$_Q$ [of$_R$ journalists$_C$]$_{P+A}$''
	\item
	``millions$_Q$ [of$_R$ dollars$_C$]$_C$''
	\item 
	``plenty$_Q$ [of$_R$ fish$_C$]$_C$''
	\item 
    ``books$_C$ [about$_R$ the$_F$ War$_P$]$_E$''
    \item
    ``People$_C$ [with$_R$ red hair]$_E$''
    \item 
    ``Words$_C$ [in$_R$ English$_C$]$_E$''
    \item 
    ``Garden$_C$ [with$_R$ trees$_C$]$_E$''  
    
\end {itemize}
\end{enumerate}

\item
{\bf Linkers}: where prepositions link two Scenes (e.g., temporal linkage, purposive linkage etc.), there are marked Ls:

\begin{itemize}
\item
  ``After$_L$ graduation$_H$, [John moved to NYC]$_H$''
\item
  ``[Due to]$_L$ [John's illness]$_H$, [the meeting is postponed]$_H$''
\end{itemize}

\item
{\bf Phrasal verbs:} the preposition changes the semantics of the verb in an unpredictable way. In that case the preposition is considered to be a part of the S or P. The P/S together form an unanalyzable unit (as it does not have sub-parts with significant semantic input).

\begin{itemize}
\item
``John$_A$ [gave up]$_P$ [his pension]$_A$''
\item
``John$_A$ let$_{P-}$ Mary$_A$ down$_{-P}$''
\item
``John$_A$ [took]$_{P-}$ Mary$_A$ [up on]$_{-P}$ [her$_A$ promise$_P$]$_A$'' 
\end{itemize}

\highlight{Note that this case does not cover cases where the preposition doesn't change the semantics of the main relation, but is mandatory, such as in ``John referred$_P$ me$_A$ [to$_R$ Mary$_C$]$_A$'', ``John relies [on$_R$ Mary$_C$]$_A$''. }

\item
{\bf Main relations:} If the preposition is the main relation in the Scene, it is an S. 

\begin{itemize}
\item
``[The apple tree]$_A$ is$_F$ in$_S$ [the garden]$_A$''
\item
``John$_A$ is$_F$ into$_S$ Mary$_A$''
\end{itemize}

\end{enumerate}

\section{Classification of Possessives}\label{sec:possessives}

Possessive constructions are used to indicate a relationship (of ownership but not only) between the possessor (dependent noun) and the possessee (head noun). 

For example in ``John's cat'', John is the possessor, cat is the possessee and the possessive marker 's indicates the relationship between them. In ``chairman of the parliament'', ``chairman'' is the possessee and ``the parliament'' is the possessor.

We distinguish between the following cases: 

\begin {enumerate}
\item
If the head noun (the possessee) is scene evoking: 
We will mark it P or S and the dependent noun will typically be marked A. 

\begin{itemize}
\item 
``[John$_C$ 's$_R$]$_A$ graduation$_P$ [from college]$_A$''
\item
``John$_A$ is$_F$ the$_F$ father$_S$ [of$_R$ that$_E$ child$_C$]$_A$'' (relational nouns including kinship terms are scene evoking) 
\item 
``[My$_A$ friend$_{S+A}$]$_A$ is$_F$ relocating$_P$'' \footnote{In principle ``my'' could be A+R. We omit R because it's a residual category (see Section~\ref{sec:technical_notes}).}
\item 
``Mary liked [[John$_C$ 's$_R$]$_A$ idea$_S$]$_A$''
\end{itemize}

\item
Ownership (including of pets) evokes a scene where the possessive construction itself is marked S: 

\begin{itemize}
\item
``[John$_A$ 's$_S$ \rem{car}$_A$]$_E$ car$_C$]$_A$ is parked outside''
\item
``[my$_{S+A}$ \rem{car}$_A$]$_E$ car$_C$]$_A$ is parked outside''
\item 
``[This dog]$_A$ is$_F$ mine$_{S+A}$''
\item
``[[John$_A$ 's$_S$ \rem{dog}$_A$]$_E$ dog$_C$]$_A$ runs$_P$ fast$_D$''
\end{itemize}

\item
 Social/organizational relationships are scene-evoking as well. If the head noun is not in itself scene evoking, we mark the possessive construction as the S: 

\begin{itemize}
  \item
  ``[[John$_A$ 's$_S$ \rem{house}$_A$]$_E$ house$_C$]$_A$ is near$_S$ [[his$_{S+A}$ \rem{school}$_A$]$_E$ school$_C$]$_A$'' (``John's house'' is a regular possessive construction whereas ``his school'' is social/organizational) 
\end{itemize}

\end {enumerate}

The following types of relationship do not evoke a scene:

\begin{itemize}
\item 
Inalienable possession (e.g.~body parts, one's name) 

\begin{itemize}
\item 
``John$_E$ 's$_R$ hand$_C$'' 
\item
``His$_E$ name$_C$''
\item 
``[the$_F$ car$_C$ 's$_R$]$_E$ windshield$_C$'', ``the$_F$ windshield$_C$ [of$_R$ the$_F$ car$_C$]$_E$''
\item
``Queen$_C$ [of$_R$ England$_C$]$_E$''
\end{itemize}

 
\item
Several other relations are often expressed in English with the possessive construction, despite not being much related to possession.

For instance, relational parts:

\begin{itemize}
\item
``bottom$_C$ [of$_R$ the$_F$ sea$_C$]$_C$'', ``tip$_C$ [of$_R$ the$_F$ iceberg$_C$]$_C$''
\end{itemize}

The following cases are annotated with the possessed as Q, and the possessor as C:

\begin{itemize}
\item
Quantities: ``scores$_Q$ [of$_R$ owls$_C$]$_C$'', ``[a$_F$ number$_C$]$_Q$ [of$_R$ bees$_C$]$_C$''
\item
Portions: ``some$_Q$ [of$_R$ the$_F$ cats$_C$]$_C$'', ``80\%$_Q$ [of$_R$ women$_C$]$_C$'', ``the$_F$ rest$_Q$ [of$_R$ the$_F$ cake$_C$]$_C$''
\item
Unitizers: ``box$_Q$ [of$_R$ chocolates$_C$]$_C$'', ``bottle$_Q$ [of$_R$ whiskey$_C$]$_C$'', ``[four$_Q$ episodes$_C$]$_Q$ [of$_R$ Dallas$_C$]$_C$''
\end {itemize}

Where the possessed indicates a type, it is annotated as E, and the possessor as C:

\begin{itemize}
\item
``type$_E$ [of$_R$ guy$_C$]$_C$''
\item
``this$_E$ breed$_E$ [of$_R$ horse$_C$]$_C$''
\item
``that$_E$ kind$_E$ [of$_R$ teacher$_{P+A}$]$_C$''
\end{itemize}

Where the possessed indicates an evaluation of the possessor (e.g., ``gem of a''), we annotate
the possessed as E (if the possessor is a non-scene) or a D (if the possessor is a P/S).

\begin{itemize}
\item
Evaluation: ``gem$_E$ [of$_R$ a$_F$ person$_C$]$_C$'', 
``gem$_D$ [of$_R$ a$_F$ party$_C$]$_P$''
\end{itemize}

\end{itemize}

\paragraph{Verbs of Possession.} Verbs can be used to express possession (e.g., ``have'', ``own'' or ``possess''). Whenever a verb  carries the semantic meaning of ownership  and precedes a concrete object (e.g. book, pen), it should be marked an S.

\be 
\item 
``John$_A$ has$_S$ [a book]$_A$'' 
\item 
``John$_A$ owns$_S$ [a car]$_A$'' 
\ee

But whenever ``have'' does not evoke a Scene in itself, but is a part of a larger phrase evoking an activity or state,  then it should be marked as an F. 

\be
\item
``John$_A$ [had$_F$ a$_F$ walk$_C$]$_P$ yesterday$_T$''
\item
``John$_A$ [has$_F$ problems$_C$]$_S$''
\item
``John$_A$ [has$_F$ hobbies$_C$]$_S$''
\item
``John$_A$ [has$_F$ hobbies$_C$]$_S$''
\item
``John$_A$ [has$_F$]$_{S-}$ no$_D$ [siblings$_C$]$_{-S}$''
\ee


\section{Classification of Static Scenes and Verbless/Copula Clauses}\label{sec:copula_verbless}


In some languages, clauses can completely lack verbs (e.g., Hebrew), while in others they would minimally include a copula (e.g., English). Treatment of both cases is similar in UCCA.

\begin {enumerate}

\item
{\bf Major Types of Static Scenes.}

\begin{enumerate}
\item 
In cases where the only purpose of a scene is to state that two entities are one and the same, we mark the copular verb ``be'' as the S. 
Examples include presentational scenes (e.g ``this is a car'') and identity cases (e.g ``he is John'').

\begin{itemize}
\item 
``[The person over there]$_A$ is$_S$ John$_A$''
\item
``[The morning star]$_A$ is$_S$ [the evening star]$_A$''
\item
``This$_A$ is$_S$ [a$_F$ car$_C$]$_A$''
\item
``This$_A$ is$_S$ [an$_F$ [amazing$_S$ \rem{restaurant}$_A$]$_E$ restaurant$_C$]$_A$''
\item
``This$_A$ is$_S$ [a$_F$ [red$_S$ \rem{car}$_A$]$_E$ car$_C$]''
\end{itemize}

\item
On the other hand, whenever we use verbless/copular clauses to attribute some property to an entity,  then the attribution itself should be the S:

\begin{itemize}
\item 
``[This$_E$ car$_C$]$_A$ is$_F$ red$_S$''
\item
``John$_A$ is$_F$ tall$_S$''
\item
``he$_A$ is$_F$ old$_S$''
\item
``he$_A$ is$_F$ American$_S$'' (borderline case)
\item
``he$_A$ is$_F$ [nine$_Q$ years$_C$]$_T$ old$_S$''
\item
\hebrew{``hu$_A$ kanadi$_S$''} (``he (is) Canadian'')
\end{itemize}

\item
Verbless/copular clauses can also express a relation between two participants. In these cases, there must be another word to express the relation between the entities. That word will be marked S.

\begin{itemize}
\item 
``Shakespeare$_A$ is$_F$ the$_F$ author$_S$ [of$_R$ Hamlet$_C$]$_A$''
\item
``John$_A$ is$_F$ a$_F$ member$_S$ [of$_R$ the$_F$ NRA$_C$]$_A$''
\item
``Mary$_A$ is$_F$ my$_A$ sister$_S$''
\item
``I$_A$ am$_F$ with$_S$ John$_A$''
\end{itemize}
\end{enumerate}


\item
{\bf Change-of-state verbs} (become, go, get, turn, grow) are treated as Ds.
The same goes for verbs that that express the absence of a change of state (e.g., stay, remain, keep).

\begin{itemize}
\item
``John$_A$ grew$_D$ old$_S$''
\item
``Mary$_A$ turned$_D$ ill$_S$''
\item
``John$_A$ stayed$_D$ awake$_S$''
\end{itemize}

\item {\label{seem}}
{\bf Verbs of perception and sense:} verbs such as seem/look/appear/sound/feel are often used without specifying the experiencer of the feeling/perception. In these cases they should be treated as a G. In case where the experiencer is stated, they should be a separate Scene.

\begin{itemize}
\item 
``[The coffee]$_A$ seems$_G$ to$_F$ be$_F$ hot$_S$''
\item
``[The car]$_A$ looks$_G$ good$_S$''
\item
``It$_F$ seemed$_P$ [to$_R$ Mary$_C$]$_A$ [that$_R$ [the coffee]$_A$ is$_F$ hot$_S$]$_A$''
\item
``It$_F$ appears$_G$ that$_F$ he$_A$ had$_F$ left$_P$ [the country]$_A$''
\end{itemize}

\item
{\bf Other types of static scenes}

\begin{enumerate}



\item
Benefaction: 

\begin{itemize} \item``[This$_F$ present$_C$]$_A$ is$_F$ for$_S$ [John]$_A$'' \end{itemize}

\item
Location: 
\begin{itemize} \item``[The$_F$ apple$_E$ tree$_C$]$_A$ is$_F$ in$_S$ [the$_F$ garden$_C$]$_A$'' \end{itemize}

\item 
Ownership (except for cases of possession used to express a body part, e.g., ``my hand'', 
which is not a Scene):

\begin{itemize}
\item 
``[This$_E$ book$_C$]$_A$ is$_F$ John$_A$ 's$_S$''
\item
``[This$_E$ book$_C$]$_A$ is$_F$ mine$_{S+A}$'' (S+A: both an S and an A)
\item
``[[my$_{S+A}$ \rem{book}$_A$]$_E$ book$_C$]$_A$ is$_F$ red$_S$''

\end{itemize}
\end{enumerate}








\item
\textbf{Existential ``There''.} 

We mark existential ``There'' as the State unless:
\begin{itemize}
    \item 
    there is a locative phrase after the entity, or
    \item
    the thing that exists is scene-evoking
\end{itemize}

Examples:
\begin{enumerate}
    \item 
    Where existential ``There'' is marked as State:
    \begin{itemize}
        \item 
        ``There$_S$ are$_F$ [thousands$_Q$ [of$_R$ us$_C$]$_C$]$_A$''
        \item
        ``There$_S$ is$_F$ [a$_F$ restaurant$_C$]$_A$''
        \item
        ``There$_S$ is$_F$ [a$_F$ [great$_S$ \rem{restaurant}$_A$]$_E$ restaurant$_C$]$_A$''
        \item
        ``There$_S$ is$_F$ [a$_F$ man$_C$ [who$_R$ can$_D$ solve$_P$ this$_A$ \rem{man}$_A$]$_E$]$_A$''
        
    \end{itemize}
\item
Where there is a locative phrase after the entity:
\begin{itemize}
    \item 
    ``There$_F$ are$_F$ earrings$_A$ on$_S$ [the$_F$ table$_C$]$_A$''
    \item
    ``There$_F$ is$_F$ [a$_F$ restaurant$_C$]$_A$ nearby$_{S+A}$''
    \item
    ``There$_F$ is$_F$ [a$_F$ [great$_S$ \rem{restaurant}$_A$]$_E$ restaurant$_C$]$_A$ in$_S$ [the$_F$ park$_C$]$_A$''
\end{itemize}
\item
 Where the thing that exists is scene evoking:
 \begin{itemize}
\item
``There$_F$ is$_F$ [a$_F$ wedding$_C$]$_P$ [in$_R$ the$_F$ park$_C$]$_A$''
\end{itemize}
\end{enumerate}

\noindent
\textbf
{Note:} the category is not defined by the words comprising the unit, but by the function the unit has in the unit it is placed in.
Consider these pairs of examples:
\be
\item
``John$_A$ is$_F$ sitting$_S$ [in$_R$ the$_F$ garden$_C$]$_A$'' / ``[The$_F$ apple$_E$ tree$_C$]$_A$ is$_F$ in$_S$ [the$_F$ garden$_C$]$_A$'' 
\item
``[John$_A$ bought$_P$ wine$_A$]$_H$ for$_L$ [[Mary$_C$ 's$_R$]$_A$ birthday$_P$]$_H$''
/ ``[This present]$_A$ is$_F$ for$_S$ [[John$_C$ 's$_R$]$_A$ birthday$_P$]$_A$''
\ee
\end{enumerate}









\section{Participant-Adverbial Distinction}\label{app:AD-distinction}

\begin{itemize}
\item
Any unit that introduces a new participant is an A. Subjects, objects, instruments, locations,
destinations are therefore invariably As.
\item
Adverbs and any other units that introduce another relation (without introducing a participant) 
into the Scene are Ds. Manner adverbs (e.g., ``quickly'', ``politely'') are invariably Ds.
\item
Prepositional phrases constitute most of the borderline cases.
\end{itemize}

\noindent
{\bf Examples:}

\begin{enumerate}
\item
``John$_A$ suffered$_P$ [for$_R$ the$_F$ team$_C$]$_A$''
\item
``Woody$_A$ walked$_P$ [in$_R$ the$_F$ park$_C$]$_A$ yesterday$_T$''
\item
``John$_A$ cut$_P$ [the$_F$ cake$_C$]$_A$ [with$_R$ a$_F$ knife$_C$]$_A$''
\item
``John$_A$ behaved$_P$ recklessly$_D$''
\item
``Woody$_A$ treated$_P$ him$_A$ [with$_R$ disrespect$_C$]$_D$''
\item
``Texas$_A$ won$_P$ [in$_R$ its$_E$ home$_E$ court$_C$]$_A$''
\item
``John$_A$ bought$_P$ milk$_A$ [next door]$_A$ [for$_R$ 50$_Q$ p$_C$]$_A$'' (``next door'' is a location, albeit a vague one)
\end{enumerate}

\section{Analyzability}

By default, analyze all cases down to the word level.
But if there is a word whose meaning is specific to the context of an expression (e.g., a person's name) or 
it's not clear how that word contributes meaning to the expression, the expression is unanalyzable.\\[3pt]


\noindent \textbf{Unanalyzable expressions include:}

\begin{itemize}
\item Personal names
  \begin{itemize}
  \item
  ``Dr.$_E$ [John Q.~Smith UNA]$_C$'': Given/family names form an unanalyzable unit, though E applies to honorific titles like ``Dr.'', ``Saint'', ``President'', ``Queen''.
  \end{itemize}
\item Titles of works of art/literature/law: ``A Tale of Two Cities'' (book), ``Marbury v.~Madison'' (legal case)
\item Idiomatic multiword expressions with opaque meaning
  \begin{itemize}
  \item ``hot dog'' (food), ``give up'' (`quit'), ``the real deal'', ``kick the bucket'' (`die'), ``in order to'', ``as well as'' (`and also')
  \item This includes phrases from another language: ``cr\`{e}me de la cr\`{e}me'' (in English)
  \end{itemize}
\end{itemize}

\noindent Proper names of places, organizations, and events are generally \textbf{analyzable}, as are many specialized/technical terms:

\begin{itemize}
\item
``Silicon$_E$ Valley$_C$''
\item
``Microsoft$_C$ Corporation$_E$'', ``University$_C$ [of$_R$ California$_C$]$_E$'', ``UC$_C$ Berkeley$_C$'', ``Society$_C$ [of$_R$ [Linguistics$_E$ Under\-graduate$_E$ Students$_C$]$_C$]$_E$'', ``[Food$_C$ and$_N$ Drug$_C$]$_E$ Administration$_C$'' 
\item
``World$_E$ War$_C$ II$_Q$'', ``the October$_E$ [Revolution$_P$]$_C$'' (although this refers to a specific revolution)
\item
``chief$_E$ executive$_E$ officer$_C$''
\item
``[natural$_E$ language$_C$]$_A$ processing$_P$''
\item
``time$_E$ signature$_C$'' (music)
\item
``French$_E$ horn$_C$'' (kind of musical instrument whose design is historically associated with France). Although this expression's meaning is not fully predictable from its parts, it is possible to recognize the semantic input\slash contribution of all parts, hence it is analyzable.
\end{itemize}

\noindent If something is named after a different kind of thing (e.g.~city named after person), that thing is not analyzed internally. Compare:
\begin{itemize}
\item
``[St.$_E$ Lawrence$_C$]$_A$ was a kind man''
\item
``I live [by$_R$ the$_F$ [St.~Lawrence UNA]$_C$ River$_E$]$_A$''
\item
``I live [in$_R$ [St.~Paul UNA]$_C$]$_A$''
\end{itemize}

\section{Detailed Guidelines}

\subsection{Scenes}

\highlight{\paragraph{Annotating Scenes within Scenes.} 
In order to analyze a Scene within a Scene we have two options: 

\begin{enumerate}
\item Analyze it first with Center-Elaborator relations (see Section~\ref{model2} for elaboration on non-Scene units).
\item Analyze it first with Process/State-Participant relations (see Section~\ref{model1} for elaboration on Scene units). 
\end {enumerate}

To determine this we ask ourselves what would we mark as the Center? If it's a concrete entity then we begin with Center-Elaborator relations, but if it's some kind of action or state then we annotate it directly as a Scene.

\begin{itemize}
\item Analysis of a Scene within Scene first with Center-Elaborator relations: 
\begin{itemize}
\item
``[The$_F$ dog$_C$ [that$_R$ ate$_P$ [my homework]$_A$ \rem{dog}$_A$]$_E$ ]$_A$ is$_F$ brown$_S$''
\item
``I$_A$ like$_S$ [[burned$_S$ \rem{coffee}$_A$]$_E$ coffee$_C$]$_A$''
\item
``Brad$_A$ played$_P$ [an$_F$ American$_C$ [\rem{American}$_A$ going$_P$ [to$_R$ the$_F$ Adriatic$_C$ ]$_A$ ]$_E$ ]$_A$''
\end{itemize}

\item Analysis of a Scene within Scene directly as a Scene: 

\begin{itemize}
\item 
``[[John$_C$ 's$_R$]$_A$ kick$_P$] saved$_P$ [the$_F$ game$_P$]$_A$'' 
\item
``John$_A$ said$_P$ [he$_A$ invented$_P$ [skating$_P$]$_A$]$_A$'' 
\end{itemize}

\end{itemize}}

\paragraph{Dependent Scenes.}
A Scene is not necessarily something that can stand on its own. It may require
a larger construction to rely on, but it is still considered a Scene:
\be
\item
``[he$_A$ retired$_P$]$_H$ [with]$_L$ [ [ [a$_F$ rank$_C$]$_S$ [of$_R$ major$_C$]$_A$ \rem{he}$_A$ ]$_H$''
\item
``[once$_T$ poor$_S$ \rem{he}$_A$]$_H$, [he$_A$ now$_T$ owns$_S$ [a$_F$ [spacious$_S$ \rem{apartment}$_A$]$_E$ apartment$_C$]$_A$ ]$_H$''
\ee

\paragraph{Distinguishing Ground and Participant Scenes.}
A Ground unit relates to the speech event itself or some aspect of it. 
It does not introduce a new Scene above and beyond referring to the speech event.
We do not analyze the internal structure of Gs.
\be
\item
``[ [The truth is that]$_G$ John$_A$ is$_F$ [a$_F$ conservative$_C$]$_S$]$_H$''
\item
``[ Surprisingly$_G$ , I$_A$ saw$_P$ John$_A$ [in the park]$_A$]$_H$''
\item
``[ [To my surprise]$_G$ I$_A$ saw$_P$ John$_A$ [in the park]$_A$]$_H$''
\item
``[you$_A$ can$_D$ go$_P$ home$_A$, [for all I care]$_G$ ]$_H$''
\ee

Contrast with participant Scenes. Both ``I think'' and ``Mary saw'' introduce a new Scene, with a new P. They are therefore participant Scenes.
\be \item
``I$_A$ think$_P$ [that$_R$ John$_A$ is$_F$ [a$_F$ conservative$_C$]$_S$ ]$_A$''
\item
``Mary$_A$ saw$_P$ [John$_A$ running$_P$ [in$_R$ the$_F$ park$_C$]$_A$ ]$_A$''
\ee

\paragraph {Speaker attitude  - distinguishing between Ground, Adverbial and State.}

\begin {enumerate}
\item
Any unit that relates to a certain quality of the main event in a Scene should be marked D.

\begin {itemize}
\item
``We had an amazing$_D$ [test drive]$_P$ !''
\item
``He ran$_P$ amazingly$_D$ !''
\end {itemize}

\item
Any unit that describes a certain quality of a concrete entity in a Scene should be marked S and the entity -- A. 
\begin {itemize}
\item
``Amazing$_S$ book$_A$!''
\end {itemize}

\item
Any unit that expresses the speaker's attitude toward the event, but doesn't directly describe a certain 
quality of the P/S should be marked G:

\begin{itemize}
\item 
``Amazingly$_G$, we had an excellent time'' (We would have had the same excellent time even if the speaker wasn't amazed)
\item
``Surprisingly$_G$ he went there''
\item
``Interestingly$_G$, he decided to do it'' 
\item 
``They shockingly$_G$ decided to get a divorce''
\end{itemize}
\end {enumerate}

\paragraph{Scene or not a Scene.} One of the most important decisions in UCCA annotation is to determine whether a relation is an S/P (and evokes a Scene) or not. 
Processes are usually easier to spot -- they describe an event that evolves in time, usually some action or movement. As for States, they differ from non-Scenes 
in not being inherent properties of the Center, but something that may have been different in the past or will be different in the future.

\be
\item
``[The$_F$ outbreak$_C$]$_D$ [of$_R$ the$_F$ War$_C$ ]$_P$'' -- a Scene.
\item
``Oscillating$_P$ [between$_R$ atheism$_C$ and$_N$ agnosticism$_C$]$_A$'' -- a Scene.
\item
``[John$_C$ 's$_R$]$_A$ accurate$_D$ kick$_P$'' -- a Scene.
\item
``[[broken$_S$ \rem{glass}$_A$]$_E$ glass$_C$]$_A$ is$_F$ dangerous$_S$'' -- a Scene.
\item
``John$_A$ always$_D$ wanted$_P$ [a$_F$ garden$_C$ [with$_R$ trees$_C$]$_E$]$_A$'' 
\item
``The trees are$_F$ in$_S$ [the$_F$ garden$_C$]$_A$'' -- a Scene, since being in the garden is not an inherent property of the trees.
\ee

\paragraph{One Scene or two.} Where two potentially Scene-evoking relations appear in proximity to one another, 
the question of whether to consider them one complex Scene or two separate ones arises. 
It should be one Scene if the two relations are conceptually hard to separate and are similar in their participants, time, location and ground. 
It should be two Scenes if this is not the case.

\be
\item
  ``[I got home]$_H$ and$_L$ [took a shower]$_H$'' (2 Scenes with a temporal relation)
\item
  ``[it took a lot of effort]$_H$ to$_L$ [win this fight]$_H$'' (2 Scenes, with a purposive relation)
\item
  ``[he is on vacation]$_H$, [sailing a yacht near Greece]$_H$'' (2 Scenes)
\item
  ``[John$_A$ eats$_P$ \rem{enthusiastically}$_D$]$_H$ and [drinks$_P$ enthusiastically$_D$ \rem{John}$_A$]$_H$'' (2 Scenes, ``eating'' and ``drinking'' are two conceptually different actions)
\item
``[She$_A$ [went away]$_P$]$_H$ [angry$_P$ \rem{She}$_A$]$_{H}$'' 
\ee

\paragraph{Secondary Verb or Participant Scene.} Distinguishing between secondary verb constructions and Participant Scene constructions is done by determining whether the sentence in question refers to one or two Scenes. 
Participant Scenes correspond to cases where there are two separate Scenes, while secondary verbs correspond to the cases where there are two relations, 
one dependent (secondary, not evoking a Scene in its own right, could not by itself be the P/S of a Scene) 
and one independent (the main relation) within the same Scene.

\be
\item
``He$_A$ demanded$_P$ [to$_F$ see$_P$ [the$_F$ manager$_{S+A}$]$_A$ \rem{He}$_A$]$_A$]'' (2 Scenes, since the demanding and the seeing are two separate Scenes which can take place in different times and locations)
\item
``He$_A$ began$_D$ kicking$_P$ [the ball]$_A$'' (one Scene, since ``began'' does not describe an action in its own right, but is dependent on the ``kicking'')
\item
``He$_A$ wants$_D$ to$_F$ kick$_P$ [the ball]$_A$'' (one Scene, since ``wants'' does not describe an action in its own right, but is dependent on the ``kicking'')
\item
  ``He$_A$ became$_D$ [a$_F$ doctor$_C$]$_S$'' (one Scene; the becoming and him being a doctor are the same conceptual event)
\item
``He$_A$ is$_F$ known$_P$ [as$_R$ [a$_F$ doctor$_C$]$_S$ \rem{He}$_A$ ]$_A$'' (two Scenes; him being known to be something and him being a doctor)

\item
  ``John$_A$ said$_P$ [he$_A$ is$_F$ [a$_F$ doctor$_C$]$_S$]$_A$'' (two Scenes; John saying and him being a doctor are easy to conceptualize as two different scenes, the sentence just places them together)
 
\ee

\paragraph {Scene within Scene or two Parallel Scenes.} In order to decide whether a Scene should be included within a larger Scene we first need to ask what role it will be assigned.
If we think it is an A then we can indeed include it as an A-Scene in the larger Scene. But If we think it should be a D or T then we instead mark it separately as an H since Ds and Ts cannot be Scenes. 

\begin {itemize}
\item 
``John said [that two men were  fighting in the street]$_A$'': Scene within Scene  (``men fighting in the street'' is an A Scene in the larger Scene).
\item 
``[John usually plays soccer]$_H$ after$_L$ [he finishes his homework]$_H$'': two Parallel Scenes (if we replace ``he \ldots homework'' with a simple non-Scene unit, e.g., ``John usually plays soccer after 16:00'', then it's clear that the relation between the units is Time, but since T can't be a Scene, we mark it as an H instead).
\item 
``[You didn't do it]$_H$ [the way]$_L$ [you should have \rem{do}$_P$ \rem{it}$_A$]$_H$'' two Parallel Scenes
(``The way you should have'' relates to the manner in which ``you didn't do it'', and therefore can theoretically be referred to as a D, but since Ds can't be Scenes we mark it as an H.)
\end {itemize}

\paragraph{Verbs that can be primary or secondary.} 
There are certain verbs that in some cases will function as secondary verbs (and therefore as Ds) and in other cases as primary verbs and this depends on the context in the specific scene under question.
\be
\item
``John$_A$ remembered$_D$ to$_F$ take$_P$ [the keys]$_A$'' (context-dependent, but it's very likely that the ``remembered'' here is mostly for emphasis and therefore secondary)
\item
``John$_A$ remembered$_P$ [ [the$_F$ hike$_C$]$_P$ [with all his friends]$_A$ ]$_A$''
\item
``John$_A$ forgot$_P$ [ how$_C$ [\rem{how}$_D$ to$_F$ ride$_P$ [his bicycle]$_A$]$_E$ ]$_A$'' (clearly the forgetting and the riding are not in the same time) 
\ee

\paragraph{Infinitive clauses.} An infinitive verb can be

\begin{itemize}
\item A semantically primary verb, syntactically a complement of the secondary verb:
	\be
	\item ``He wanted$_D$ to$_F$ come$_P$ home$_A$'' 
    \item ``He wanted$_D$ me$_A$ to$_F$ come$_P$ home$_A$'' 
    \ee

\item 
A P/S of a Scene which serves as a participant in a larger scene:
	\be
	\item ``[to$_F$ kick$_P$ [a penalty shot]$_A$ [in soccer]$_A$ \imp$_A$]$_A$ is$_F$ exciting$_S$''
    \item ``John$_A$ promised$_P$ [to$_F$ be$_F$ better$_S$ \rem{John}$_A$]$_A$''
	\item ``John$_A$ is$_F$ likely$_D$ to$_F$ leave$_P$'' 
    \item ``It$_F$ is$_F$ likely$_D$ to$_F$ rain$_P$''
    \ee
\item A P/S for a scene which elaborates a non-scene unit:
	\be
    \item ``a$_F$ couch$_C$ [\imp$_A$ to$_F$ sleep$_P$ [on$_R$ \rem{couch}$_C$]$_A$]$_E$''
    \ee
\item The state or process for a scene which is parallel to another (``to'' here serves as a purposive linker): 
	\be
    \item ``To$_L$ [win$_P$ \rem{you}$_A$]$_H$, [you$_A$ [have to UNA]$_D$ find$_P$ [the$_F$ key$_C$]$_A$]$_H$''
    \item ``[a$_F$ procedure$_P$]$_H$ to$_L$ [\imp$_A$ ensure$_P$ quality$_A$]$_H$''
    \ee
\end{itemize}



By convention, when ``to'' is used as an F (same for ``zu'' in German), it should not be included within the process/state.
``For'' can introduce the subject of an infinitive clause: 

\be
\item ``I paid [[for$_R$ you$_C$]$_A$ to$_F$ eat$_P$ dinner$_A$]$_A$''
\item ``a$_F$ couch$_C$ [[for$_R$ people$_C$]$_A$ to$_F$ sleep$_P$ [on$_R$ \rem{couch}$_C$]$_A$]$_E$''
\item ``[a$_F$ procedure$_P$]$_H$ [[for$_R$ people$_C$]$_A$]$_{H-}$ to$_L$ [ensure$_P$ quality$_A$]$_{-H}$'' 
\ee

\paragraph {Relative clauses.} 

\begin {enumerate}
\item
If the RC modifies a noun that serves as a scene element (say A) then the RC should be included as an E-scene:

\begin{itemize}
\item 
This is true for nouns that are not scene-evoking like software: ``[They$_A$ tested$_P$ me$_A$ [on$_R$ the$_F$ software$_C$ [I included on my resume \rem{software}$_A$]$_E$]$_A$]$_H$''
\item
It is also true for roles that evoke P+A or S+A like judge: ``[[The$_F$ [judge$_{P+A}$]$_C$ [who$_R$ swims$_P$ \rem{judge}$_A$]$_E$]$_A$ reached for the book]$_H$''
\item
It is also true for scene-evoking nouns like wedding: ``[[The$_F$ [wedding$_P$]$_C$ [I$_A$ went$_P$ [to$_R$ \rem{wedding}$_P$]$_A$]$_E$]$_A$ was$_F$ attended$_P$ [by famous people]$_A$]$_H$''
    
\end{itemize}

\item    
On the other hand, where an RC modifies a noun which serves as the P or S of a top-level (H) scene then we have no other option than to separate the RC into a parallel scene:
    
\begin{itemize}
\item
``[I$_A$ [have$_F$ a$_F$ friend$_C$]$_{S+A}$]$_H$ who$_L$ [\rem{friend}$_A$ drove$_P$ all the way from Tel-Aviv]$_H$ to$_L$ [\rem{friend}$_A$ get$_P$ here$_A$]$_H$''
\item
``[The$_F$ party$_P$ ]$_{H-}$ [I$_A$ went$_P$ [to$_R$ \rem{party}$_P$]$_A$]$_H$ [was$_F$ fun$_D$]$_{-H}$''
\end{itemize}
\end{enumerate}

\paragraph{Secondary predicates.} 
A depictive or resultative should be marked separately from the main predicate as an independent parallel Scene.

\begin {enumerate}
\item 
Depictives: 

\begin{itemize}
\item
``[John$_A$ left$_P$ home$_A$]$_H$ [young$_S$ \rem{John}$_A$]$_H$''
\item 
``[John$_A$ ate$_P$ [the food]$_A$]$_H$ [cold$_S$ \rem{food}$_A$]$_H$''
\item
``[He$_A$ left$_P$ [the party]$_A$]$_H$ [angry$_S$ \rem{he}$_A$]$_H$''
\end{itemize}

\item
Resultatives: 

\begin {itemize} 
\item
``[Mary$_A$ painted$_P$ [the fence]$_A$]$_H$ [blue$_S$ \rem{fence}$_A$]$_H$''
\item
``[He$_A$ [cried himself]$_P$]$_H$ to$_L$ [sleep$_P$ \rem{He}$_A$]$_H$''
\end {itemize} 
\end {enumerate}



\highlight{\paragraph{Cognitive events.} Cognitive events (e.g., thinking, seeing, wondering) should be marked as Processes. 

\be 
\item 
``I$_A$ see$_P$ [that you both are getting along]$_A$''
\item
``I$_A$ think$_P$ [it's OK]$_A$''
\item
``I$_A$ wonder$_P$ [whether we're doing a mistake]$_A$''
\ee}

\paragraph{Results of Scenes.} Results of Scenes can be Scenes in their own right. 

\be
\item
``[the$_F$ outcome$_C$]$_P$ [[of$_R$ the$_F$ trial$_C$]$_P$ ]$_A$''
\ee


\paragraph{Noun Scenes.} A noun Scene is a case when a noun-phrase serves as a Scene and the noun itself is the main relation in the Scene (the P or S). 
They should be internally analyzed as Scenes, with a P/S, As, Ds and Ts. 
In general, deverbal nouns are cases of noun Scenes, although not all noun Scenes are formed by deverbal nouns. 

\be
\item
``[[John$_C$ 's$_R$]$_A$ accurate$_D$ kick$_P$]$_A$ saved$_P$ [the game]$_A$''
\item
``[Him$_A$ destroying$_P$ [the city]$_A$ ]$_A$ was$_F$ [a$_F$ disaster$_C$]$_S$''
\item
``[[The$_F$ destruction$_C$]$_P$ [of the city]$_A$ ]$_A$ was$_F$ [a$_F$ disaster$_C$]$_S$''
\item
``[[His]$_A$ destruction$_P$ [of$_R$ the city]$_A$ ]$_A$ was$_F$ [a disaster]$_S$''
\item
``[Gone with the Wind]$_A$ is$_F$ one$_Q$ [of$_R$]$_{S-}$ [Selznick$_C$ 's$_R$]$_A$ [productions$_C$]$_{-S}$'' 
\item
``[War$_P$]$_A$ is$_F$ imminent$_S$'

\ee
More generally, words that derive a participant from a scene are treated as scenes.

\be
\item ``[taxi$_A$ drivers$_{P+A}$]$_A$ are$_F$ usually$_D$ old$_S$''
\item ``[participants$_{P+A}$]$_A$ are$_F$ welcome$_S$''
\ee

Nouns denoting an individual participant of a kinship relation or static social or organizational relationship (``father'', ``friend'', ``boss'', ``employee'', ``chairman'') should evoke an S scene 
of the relationship.\footnote{An exception is a case where the noun is used as a title, like ``Chairman Mao'', ``Prof. Smith'', or ``Father Smith (priest)''.}

\be
\item
``[John$_C$ 's$_R$]$_A$ father$_{S+A}$''
\item
``my$_{A}$ father$_{S+A}$''
\item
``John$_A$ [has$_F$]$_{S+A-}$ no$_D$ [father$_C$]$_{-S+A}$''
\item
``John$_A$ is$_F$ [Paul$_C$ 's$_R$]$_A$ father$_{S}$''
\ee


\paragraph{Scenes without P/S.} Some Scenes have no P or S, since it is omitted or implied. In this case, we should add them as remote/implicit units.

\be
\item
``[John bought eggs]$_H$ and$_L$ [Mary$_A$ [chewing gum]$_A$ \rem{bought}$_P$ ]$_H$''
\item
``[John$_A$ wanted$_P$ [a real life]$_A$ ]$_H$, [not$_D$ [life in a caravan]$_A$ \rem{John}$_A$ \rem{wanted}$_P$]$_H$''
\item
``[how about]$_S$ coffee$_A$?''
\ee



\paragraph{Imperatives.}\label{imperatives} Imperative clauses should be marked:
\begin{enumerate}
\item
As a Scene with an Implicit A if the party addressed does not appear explicitly in the text:
\begin {itemize}
\item
``Stop$_P$ \imp$_A$ !''
\item 
``Please$_F$ [take care]$_P$ [of your brother]$_A$ \imp$_A$''
\end{itemize}
\item
As a Scene with Remote if the addressee appears explicitly in the text:
\begin{itemize}
\item
``[\rem{You}$_A$ Eat]$_H$. [You$_A$ 'll feel better]$_H$.''
\end{itemize}
\item
If the addressee appears as a vocative, it should be included in the Scene as G+A:
\begin{itemize}
\item
``[John$_{G+A}$, go$_P$ outside$_A$]$_H$''
\end{itemize}
\end{enumerate}


\paragraph {Answer fragments.} We should add the missing elements from the preceding question.

\begin {itemize}
\item 
``Will you be able to make it?'' ``I most certainly will.''
In this case, the answer should be marked: [ I$_A$ [most$_E$ certainly$_C$]$_D$  will$_F$ \rem{able}$_D$ \rem{make it}$_P$]$_H$''
\item
``When will John come?'' ``Tomorrow''. 
Here the answer should be marked: ``[Tomorrow$_T$ \rem{John}$_A$ \rem{come}$_P$]$_H$''
\end {itemize}

As for ``yes'', ``no'' or any other affirmation/negation answer fragments: they should be marked as a G inside an H together with the implicit scene which should be added from the question.
For example:
\begin {itemize}
\item
``Is John coming? Yes.''
The answer should be marked: ``[yes$_G$ \rem{John}$_A$ \rem{coming}$_P$]$_H$''
\item
Of course, in cases where the scene is not elided (e.g ``Is John coming?'' ``Yes, John is coming'') we don't have to add the scene from the question, and can directly mark the whole answer together with the affirmation fragment as an H: 
``[yes$_G$, John$_A$ is$_F$ coming$_P$]$_H$''

Note that this solution also simplifies some technical challenges. For example: 
``Is John coming? surprisingly, yes''
Here we can mark both ``surprisingly'' and ``yes'' as separate Gs inside the H together with the implicit scene which we will add from the question:
``[surprisingly$_G$ yes$_G$ \rem{John}$_A$ \rem{coming}$_P$]$_H$''
Otherwise, if we didn't add the implicit scene from the question and just left it ``surprisingly, yes'' it would be hard to reflect the relation between ``surprisingly'' and ``yes''.
\end{itemize}

\paragraph {Thanks/Thank you.} We differentiate between two cases:

\begin {enumerate}
\item
When the Participant ``I'' is implicit:
In such cases ``thanks'' and ``thank you'' should be marked P and an Implicit A should be added to stand for the person thanking. 

\begin {itemize}
\item 
``[[Thank you  UNA]$_P$ [for your wonderful hospitality]$_A$ , \imp$_A$]$_H$''
\item
``Thanks$_P$ [for your wonderful hospitality]$_A$ \imp$_A$]$_H$]''
\item
``[[Thank you UNA]$_P$ everyone$_A$ [for coming]$_A$ \imp$_A$]$_H$''
\item
``[Thanks$_P$, John$_G$! \imp$_A$]$_H$''
\item
``[Many$_D$ thanks$_P$, \imp$_A$]$_H$''
\item 
``[Everything was absolutely great]$_H$ so$_L$ [thanks$_P$ \imp$_A$]$_H$''
\end {itemize}

\item When the person who is thanking is explicitly mentioned:
In such cases  we don't mark ``thank you'' as one phrase but mark separately ``thank'' as P and ``you'' as A.

\begin {itemize}
\item
``I$_A$ want$_F$ to$_F$ thank$_P$ you$_A$ [for coming]$_A$''
\item 
``I$_A$ would$_F$ like$_F$ to$_F$ thank$_P$ you$_A$ [for your help]$_A$.''
\end {itemize}

\end {enumerate}

\paragraph{Expletive it.} Sometimes ``it'' is used to take the place of the subject when there is no other A which does so. In this case it should be marked as an F.

\begin{itemize}
\item
``It$_F$ was$_F$ suspicious$_S$ [that$_R$ I$_A$ saw$_P$ him$_A$ there$_A$]$_A$''
\end{itemize}

\noindent
But where modality is used (e.g., ``likely'', ``unusual''), the analysis is

\begin{itemize}
\item
``It$_F$ is$_F$ likely$_D$ to$_F$ rain$_P$''
\end{itemize}

\paragraph {Cooperating participants.} If two participants cooperatively participate in the same Process or perform it in 
an identical manner then they should be united in one A with two Cs. This only applies if they are coordinated, as in ``John and Mary played tennis'', not ``John played tennis with Mary.''

\begin {itemize} 
\item
``[John$_C$ and$_N$ Mary$_C$]$_A$ went to the park''
\item
``A conversation was held [between$_R$ [[[the Prime Minister]$_{P+A}$]$_C$ and$_N$ [[the Queen]$_{P+A}$]$_C$]$_C$]$_A$''
\end {itemize} 

\paragraph{Two Types of verbs that take a participant Scene.} Note that some verbs with a participant Scene have a remote unit taken from the participant Scene or vice versa. Other verbs do not exhibit such behavior.
\be
\item
``I$_A$ expected$_P$ [John$_A$ to$_F$ come$_P$]$_A$''
\item
``We$_A$ agreed$_P$ [for John to give the funeral oration]$_A$''
\item
``I$_A$ persuaded$_P$ John$_A$ [to$_F$ come$_P$ \rem{John}$_A$]$_A$''
\item
``John$_A$ promised$_P$ [to$_F$ be$_F$ better$_S$ \rem{John}$_A$]$_A$''
\ee



\subsection{Processes/States.}

\paragraph{Modals and Auxiliaries.} 
Modals should invariably be annotated as secondary verbs (and therefore as Ds). This applies to ``would'' as well. Auxiliary verbs (``be'', ``have'', ``will'' and ``do''), which do not have significant semantic input in their own right\footnote{UCCA in its foundational layer does not annotate tense. Even if it did, the tense would not be considered a feature encoded on the auxiliaries, but rather in the combination of the auxiliary and the inflection.} are considered Fs.

\be 
\item
``John$_A$ will$_F$ come$_P$''
\item
``Mary$_A$ should$_D$ come$_P$''
\item
``Mary$_A$ is$_F$ coming$_P$''
\item
``John$_A$ [has to]$_D$ come$_P$''
\item
``I$_A$ have$_F$ done$_P$ it$_A$''
\item
``John$_A$ does$_F$ n't$_D$ know$_P$ him$_A$''
\item
``John$_A$ is$_F$ likely$_D$ to$_F$ cry$_P$''
\ee

\paragraph{Other secondary verb classes.} Other verbs classes which are treated as secondary in UCCA when used in combination with a scene evoker (i.e., not ``try the food'' or ``practice violin'', which are primary verbs):\footnote{Following Dixon's A Semantic Approach to English Grammar}

\be
\item
	BEGINNING: e.g., begin, start, finish, complete, continue (with)
\item
	TRYING, e.g., try, attempt, succeed, fail, practice
\ee

\paragraph{Secondary Verbs with an additional role.}
Some secondary verbs may introduce another role beside the roles of the main verb. An example is ``help'', ``force'' and ``permit''. Like all secondary verbs, such verbs are considered Ds. The additional participant is marked as an A in the Scene.
\be \item
``John$_A$ helped$_D$ Mary$_A$ climb$_P$ [the ladder]$_A$''
\item
``John$_A$ forced$_D$ [Mary]$_A$ to$_F$ climb$_P$ [the ladder]$_A$''
\ee

Some frequent semantic classes of verbs that introduce an additional role:

\be
\item
WANTING: e.g., want, wish (for); hope (for); need, require; expect; intend; pretend
\item
POSTPONING: e.g., postpone, delay, defer, avoid
\item
MAKING: e.g., make, force, cause, tempt; let, permit, allow, prevent, spare, ensure
\item
HELPING: e.g., help, aid, assist
\ee

\paragraph{The case of ``want''.} The verb ``want'' in English is an interesting case, as it is ambiguous between desiring a thing (``John always wanted$_P$ a garden with trees''), desiring an action (``I want$_D$ to run$_P$ a marathon'', ``He wants$_D$ to kick$_P$ the ball''), and part of a polite speech act (``I want$_F$ to thank$_P$ you for coming'').

\paragraph {Secondary main verbs.} Sometimes the Process appears as the subject of the sentence, 
where the main verb is the secondary verb. In these cases, we still mark the secondary verb as D, and the subject as the main relation.

\be
\item
``[John$_C$ 's$_R$]$_A$ career$_P$ [ended$_C$ abruptly$_E$]$_D$''
\item  
``[The$_F$ race$_C$]$_P$ began$_D$ [early$_E$ [in$_R$ the$_F$ morning$_C$]$_C$]$_T$''
\item
``His$_A$ service$_P$ was slow$_D$''
\ee




\paragraph{Light Verbs.} Cases where the verb is almost void of meaning, and most of the meaning is determined by the object. The verb is usually ``have'', ``give'', ``take'' or ``make'' (although there are other examples). Annotation: both the the light verb and the following object should be included inside the P/S. The light verb as an F and the object as a C.
\be \item
``John$_A$ [took$_F$ a$_F$ shower$_C$]$_P$''
\item
``Mary$_A$ [gave$_F$]$_{P-}$ John$_A$ [a$_F$ smile$_C$ ]$_{-P}$''
\item
``Brad$_A$ [made$_F$ a$_F$]$_{P-}$ guest$_D$ [appearance$_C$]$_{-P}$ [on$_R$ ABC$_C$]$_A$''
\ee

\paragraph{Adjective followed by a Scene:} Analyzed as a D+P construction.
\be 
\item
``John$_A$ is$_F$ easy$_D$ to$_F$ please$_P$''
\item
``John$_A$ is$_F$ likely$_D$ to$_F$ leave$_P$''
\item
``John$_A$ is$_F$ ready$_D$ to$_F$ come$_P$''
\item
``London$_A$ is$_F$ great$_D$ for$_F$ music$_P$''
\ee
 
\paragraph{Causatives.} We view the causation word (often ``make'' or ``cause'') as a secondary verb.
\be \item
``John$_A$ makes$_D$ Mary$_A$ happy$_S$''
\item
``John$_A$ inspires$_D$ interest$_P$ [in$_R$ Mary$_C$]$_A$''
\item
``We just got$_D$ [our sunroom]$_A$ built$_P$ by Patio World''
\item
``Mary had$_D$ [her hair]$_A$ done$_P$''
\ee

\noindent
In cases where the causative word links two Scenes, we treat it as a primary verb:

\be
\item
``[Fear$_S$]$_A$ always$_D$ causes$_P$ [hate$_S$]$_A$''
\ee

\paragraph{Polite Forms.} Words that only serve as part of a construction for politely addressing someone are Fs. Each word should individually be marked as F.

\begin{enumerate}
\item 
``I would$_F$ like$_F$ to thank you for all your help''
\item
  ``Could$_F$ you$_A$ help$_P$ me$_A$, please$_F$ ?''
\german{
\item
``Gehen$_P$ Sie$_F$ raus$_D$ !'' 
\item
``[[Sie und Ihr komischer Vogel]$_{G+A}$, raus$_P$] !'' ["you and your funny bird, out!"] (here ``Sie'' is part of the vocative)
\item
``Gehen$_P$ [Sie$_C$ und$_N$ Hans$_C$] raus$_D$ !''} (here ``Sie'' is part of the vocative)
\end{enumerate}

\subsection{Averbials (Secondary Relations in Scenes).}


\paragraph{Degree Modifiers.}
When a degree modifier (e.g., ``{very/quite/somewhat} warm'') applies to a P or S, then it should be marked D:

\be
	\item ``a$_F$ [very$_D$ hot$_S$ \rem{plate}$_A$]$_E$ plate$_C$''
\ee

\noindent But if it modifies a C then it should be marked E:
\be
	\item ``You$_A$ won$_P$ [quite$_E$ handily$_C$]$_D$''
	\item ``a$_F$ [very$_E$ beautiful$_C$]$_D$ wedding$_P$''
\ee

\paragraph{Negation.} Negation is considered an adverbial.
\be \item
  ``John$_A$ did$_F$ n't$_D$ touch$_P$ [the piano]$_A$''
\item
  ``[John]$_A$ is$_F$ [no]$_D$ [joker]$_P$''
\item
  \german{``Ich$_A$ trinke$_P$ keine$_D$ Milch$_A$''}
\ee

\noindent
Some pronouns and linkers express negation on a Scene. In this case, they also serve as Ds in that scene.
    
    \be
      \item
      ``Nobody$_{A+D}$ came$_P$ [to$_R$ [the$_F$ party$_C$]$_P$]$_A$''
      \item
   	``[I$_A$ left$_P$]$_H$ without$_{L}$ [eating$_P$ [[my$_S$ \rem{banana}$_A$]$_E$ banana$_C$]$_A$ \rem{I}$_A$ \rem{without}$_D$]$_H$''  
    \ee

\paragraph{D in coordination.}\label{D} Occasionally, several entities are connected by an N, where there is a D (usually a frequency, probability or temporal relation) which relates specifically to one of them. In this case, a unit with D and C children is marked as C:
\be
\item
``He$_A$ appeared$_P$ [in$_R$ [[Head of the Class]$_C$, [Freddy 's Nightmares]$_C$, [Thirtysomething]$_C$, and$_N$ [\textbf{[for a second time]$_D$} [Growing Pains]$_C$ ]$_C$ ]]$_C$]$_A$ .''
\item
``John$_A$ is$_F$ intending$_D$ to$_F$ go$_P$ [to$_R$ [Rome$_C$, Paris$_C$ and$_N$ [\textbf{perhaps$_D$} London$_C$]$_C$]$_C$ ]$_A$''.
\dout{\item
``They$_A$ treated$_P$ us$_A$ [like$_R$ people$_C$ [not$_D$ dogs$_C$ ]''
}
\ee

\paragraph{Framing of Scenes.}
Some Scenes are wrapped in a noun phrase that frames them (e.g., ``story of'', ``rumor of'', ``belief that'). In this case, the framing noun serves a separate Scene, which takes the framed Scene as a Participant.
     
     \begin{enumerate}
     \item
       ``[the$_F$ story$_C$]$_P$ [of$_R$ [a$_F$ young$_E$ girl$_C$]$_A$ sentenced$_P$ [to$_R$ death$_P$ \rem{girl}$_A$]$_A$ ]$_A$''
     \item
       ``[the$_F$ rumor$_C$]$_P$ [of$_R$ his$_A$ retirement$_P$]$_A$''
     \item
       ``[the$_F$]$_{P-}$ strange$_D$ [belief$_C$]$_{-P}$ [that$_R$ chickens$_A$ are$_F$ immortal$_S$]$_A$''
     \end{enumerate}

\subsection{Non-Scene Units.}

\paragraph{Articles.} Articles should be annotated as Fs, unless they also annotate case (e.g., ``dem'' in German), in which case they are Rs.

\be \item
 ``The$_F$ Knesset$_C$''
\item
 ``A$_F$ big brown dog$_C$''
\ee



\paragraph{Demonstratives (this, that, these, those).}

When the demonstrative relates to a noun, 
the demonstrative should be an E:

\begin{itemize}
\item 
``Put [this$_E$ book$_C$]$_A$ on the table''
\item
\hebrew{ha$_F$ kelev$_C$ ha$_F$ ze$_E$}
\end{itemize}

\noindent
On the other hand, when the demonstrative is independent we will typically mark it A:
\begin{itemize}
\item
``This$_A$ looks$_G$ good$_S$''
\item 
``This$_A$ is$_S$ [a$_F$ [great$_S$ \rem{car}$_A$]$_E$ car$_C$]$_A$''
\item
``That$_A$ was$_F$ [an$_F$]$_{P-}$ interesting$_D$ [experience$_C$]$_{-P}$''
\end{itemize}

\paragraph{Appositions.} Appositions are cases where two consecutive units are semantically parallel and refer to the same entity. If one is a proper name and the other isn't, the proper name is the C, and the other is the E.

\be
\item
  ``John$_C$, [my$_A$ history$_A$ teacher$_{P+A}$]$_E$''
\item
  ``[my$_A$ history$_A$ teacher$_{P+A}$]$_E$, John$_C$'' 
\ee

\paragraph{Extraposition.} Cases where an E does not form a contiguous stretch of text with its center. In this case, they should be marked together as a non-contiguous unit.

\be
\item
``He saw [that painting]$_{A-}$ before, [[that lovely magnificent painting]$_E$]$_{-A}$''
\item
``I met [the guy]$_{A-}$ yesterday, [[whom I first saw in the park]$_E$]$_{-A}$''
\ee

\paragraph{Fused E Scenes.} There are many constructions that resemble an E Scene construction, but don't have a clear Center they elaborate on. We still annotate them as E Scenes, marking the pronoun (if exists) as a C:

\be \item
``[ What$_C$ [\rem{What}$_A$ I$_A$ meant$_P$]$_E$ ]$_A$ was$_S$ [I want to have dinner]$_A$''
\item
``[ Any$_Q$ recipes$_C$ [she$_A$ used$_P$ \rem{recipes}$_A$]$_E$]$_A$ are$_F$ marked$_P$ [in$_R$ red$_C$]$_D$''
\item
``you$_A$ are$_F$ playing$_P$ [with$_R$ somebody$_C$ [\rem{somebody}$_A$ better$_S$ [than$_R$ you$_C$]$_A$]$_E$]$_A$''
\ee

\paragraph{Numbers and Quantifiers} are considered Qs. The question of their scope is not addressed in the current layer of the annotation. Therefore they are considered a part of the unit with the counted item.

\be
\item
``[All$_Q$ Greeks$_C$]$_A$ are$_F$ mortal$_S$''
\item
``[Two$_Q$ bananas$_C$]$_A$ are$_F$ lying$_P$ [on$_R$ the$_F$ table$_C$]$_A$''
\item
``Millions$_Q$ [of$_R$ homes$_C$]$_C$''
\ee

\noindent Sometimes the quantifier ``floats'' after the item it counts (see also Section~\ref{sec:possessives}):

\be 
\item 
``[The$_F$ Greeks$_C$ all$_Q$]$_A$ are$_F$ mortal$_S$''
\item 
``[The$_F$ Greeks$_C$]$_{A-}$ are$_F$ [all$_Q$]$_{-A}$ mortal$_S$''
\ee

\paragraph{Focus Modifiers.} Words like \textit{also}/\textit{too}, \textit{only}/\textit{just}, and \textit{even} pertain to an item's membership in a set of things under discussion. Depending on placement in the sentence, multiple interpretations may be possible. The focus modifier belongs in a unit with the item it is most closely associated with semantically---as D if part of a scene, and E otherwise:  

\be
\item
``[John$_C$ also$_E$]$_A$ likes$_S$ cats$_A$'' (John is another person who likes cats)
\item
``John$_A$ [also$_E$]$_{A-}$ likes$_S$ [cats$_C$]$_{-A}$'' (cats are another thing John likes)
\item
``John$_A$ also$_D$ likes$_S$ cats$_A$'' (another thing we know about John is that he likes cats)
\item
``There$_S$ is$_F$ [only$_E$ one piece of cake]$_A$'' 
\item
``[The supermarket]$_A$ is$_F$ just$_D$ around$_S$ [the corner]$_A$''

\ee

\noindent This holds even if the focus modifier relates entire sentences:
\be 
\item ``John promised to eat fish and also to brush his teeth'':\\
``[John promised [\rem{John}$_A$ to eat fish]$_A$]$_H$ and$_L$ [\rem{John}$_A$ also$_D$ \rem{promised}$_P$ [\rem{John}$_A$ to brush his teeth]]$_H$''
\item
``[John likes cats.]$_H$ [Also$_D$, he gets plenty of exercise.]$_H$''
\ee

\paragraph{Directions.}

We distinguish between two cases: 

\begin{enumerate}
\item 
When a directional word stands alone and is not followed by a Participant it should be marked D: 
\begin{itemize}
\item 
``The bird flew up$_D$''
\item
``She handed the ring back$_D$''
\item
``come$_P$ in$_D$''
\end{itemize}

\item
When the directional word is followed by a Participant, it should be included in it as an R:
\begin{itemize}
\item 
``We went [across$_R$ the field]$_A$''
\item
``I ran [up$_R$ the stairs]''
\item
``John is going [back$_R$ home]$_A$''
\end{itemize}

In cases where a directional word participates in a multi-worded preposition, we will mark the whole phrase as an unanalyzable R:

\begin{itemize}
\item 
``Mary walked [[out of UNA]$_R$ John's  room]$_A$''
\item
``John lives [[across from UNA]$_R$ Mary$_C$]$_A$''
\item
``John is standing [[in front of UNA]$_R$ Mary$_C$]$_A$''

\end{itemize}

It is important in such cases to distinguish between the case of a multi-worded preposition and the case of an adverbial followed by a single-word preposition:

\begin{itemize}
\item 
``John$_A$ went$_P$ away$_D$ [to$_R$ college]$_A$''
\end{itemize}

We can tell ``away to'' is not a multi-word preposition since each part in this example still makes sense when standing alone: you can say ``John went away'' and ``John went to college''. This is not the case with  ``John walked out of the room'', that cannot be reduced to ``John walked of the room''.

\end{enumerate}


\paragraph{Passive ``by''.} The ``by'' of the passive should be annotated as R.
\be \item
``He$_A$ is$_F$ scolded$_S$ [by$_R$ many$_C$]$_A$''
\ee

\paragraph{Preposition Stranding.} A construction in which a preposition is not followed by its object. 

\begin{enumerate}
\item 
Due to this construction, an A can sometimes be completely missing from a scene while its preposition is in place. In such a case we mark the preposition as an A, with an R inside of it, and add the preposition's object as a Remote:
     
     \begin{itemize}
          \item
``The$_F$ book$_C$ [I$_A$ 'm$_F$ looking$_P$ [for$_R$ \rem{book}$_C$ ]$_A$]$_E$''
\item
``[The$_F$ wedding$_P$]$_{H-}$ [I$_A$ went$_P$ [to$_R$ \rem{wedding}$_P$]$_A$]$_H$ [was$_F$ wonderful$_D$]$_{-H}$''
\end{itemize}

\item
In other cases, both the A and its preposition appear in the scene but are discontiguous, and so we should unite them under one discontiguous parent A: 
\begin{itemize}
\item
``[This$_E$ path$_C$]$_{A-}$ has$_F$ already been$_F$ walked$_P$ [on$_R$]$_{-A}$''
\item
``[What$_C$]$_{A-}$ are$_F$ you$_A$ talking$_P$ [about$_R$]$_{-A}$?''
\end{itemize}
\end{enumerate}



\subsection{Other Relations.}

\paragraph{Ordinals.}
Ordinals (e.g., ``first'', ``second'', ``last'') always relate to a state or a process relative to which they are first/second etc. They should be marked D.

\begin{itemize}
\item
``[My$_A$ first$_D$ kick$_P$]$_A$ saved$_P$ [the game]$_A$''
\item
``[The$_F$ first$_D$ king$_{P+A}$ [of$_R$ Scotland$_C$]$_A$ ]$_A$ died$_P$ [in 858]$_T$''
\item
``I$_A$ got$_P$ here$_A$ first$_D$''
\item
``I$_A$ was$_F$ [the last]$_D$ to$_F$ arrive$_P$''
\item
``I$_A$ arrived$_P$ last$_D$''

\end{itemize}

\paragraph{Repeated Actions.} Expressions that indicate the number of occurrences of a scene are Ds.

\begin{itemize}
\item
``We$_A$ talked$_P$ [three$_Q$ times$_C$]$_D$ [over the last week]$_T$''
\end{itemize}

\paragraph{Punctuation.} Not annotated in the current layer of UCCA (even commas).

\paragraph {Interjections.} short emotional utterances referring to the preceding or following text should be marked G:

\begin{itemize}
\item 
``[Ugh$_G$ !  that$_A$ is$_F$ gross$_S$]$_H$''
\item
``[Ouch$_G$ !  he$_A$ fell$_P$ [from his bike]$_A$]$_H$''
\item
``[Whoops$_G$ !  I$_A$ forgot$_D$ to$_F$ send$_P$ it$_A$]$_H$''
\item
``[Great$_G$ !  I$_A$ just$_T$ missed$_P$ [my ride back home]$_A$]$_H$''
\item
``[Great$_G$ ! I$_A$ 'm$_F$ so$_D$ happy$_S$]$_H$''
  \end{itemize}

An exception is where an adjective utterance implicitly refers to a specific place, or a specific P/S (instead of expressing emotion regarding a certain Scene as a whole). Then instead of Ground it should be analyzed as a Scene of itself: 

\begin{itemize}
\item
``Q: How was the cake ? A: [Fantastic ! $_S$ \rem{cake}$_A$]$_H$''
\end{itemize}

Since ``cake'' is explicitly mentioned in the nearby text, we add it as Remote, but in cases where the missing unit doesn't appear, we add an Implicit A and the adjective utterance should be S. For example, when a restaurant review opens with: ``Great ! '', it probably means ``great restaurant''. But if the restaurant is not mentioned anywhere in the text, we annotate it as: `Great$_S$ ! \imp$_A$''. 

\paragraph{Fillers/Discourse Markers.} When fillers (e.g., ``then'', ``well'') or discourse markers don't convey a meaning dimension that can be captured by UCCA's foundational layer, they should be marked as F. That is, if they are not (part of) Scenes, (part of) Scene elements or Linkers, they are Fs.

\begin{itemize}
\item 
``ummm$_F$ I$_A$ heard$_P$ [you$_A$ say$_P$ that$_A$]$_A$''
\item
``I$_A$ 'm$_F$ not$_D$, ah$_F$, interested$_P$''
\item
``well$_F$, this can pose a problem''
\item
``So$_F$, this is what we're going to do''
\end{itemize}

\paragraph{Linkers with a single argument.}
We also allow Ls with a single argument. This usually happens if an L relates one Scene with everything that follows/precedes it, 
without there being any particular unit that the Scene relates to. Another case where we use a single argument linker is when one of its arguments is omitted.
An example would be a paragraph that starts with ``However'' that contrasts with everything that was written in the previous paragraph.

\paragraph {Neither/ Either/ Both} are generally marked as Qs or Ds 

\begin{itemize}
\item 
``[Neither$_Q$ [of$_R$ them$_C$]$_C$]$_A$ came$_P$ to the party''
\item
``Both$_D$ lectures$_P$ were$_F$ interesting$_D$''
\item
``John would like to learn [either$_Q$ [French$_C$ or$_N$ German$_C$]$_C$]$_A$''
\item
``[Neither$_Q$ [John$_C$ nor$_N$ Mary$_C$]]$_A$ are willing to give the lecture''
\end{itemize}

An exception is when these words link two separate Scenes, in which case they are Ls: 

\begin {itemize}
\item
``[John$_A$]$_{H-}$ both$_L$ [likes$_P$, \rem{Mary}$_A$]$_{-H}$ and$_L$ [dislikes$_P$ Mary$_A$ \rem{John}$_A$]$_H$''
\item
``Either$_L$ [you go to school]$_H$ or$_L$ [you won't be allowed to play soccer today]$_H$''
\end{itemize}

\paragraph {Elaboration of/by a Coordination.} When a certain unit relates to multiple units that carry an identical role, we unify all the multiple units under one parent unit. 

\begin {itemize}
\item
``I have [10$_Q$ [brothers$_C$ and$_N$ sisters$_C$]$_C$]$_A$'' 
\item
``Queen$_C$ [of$_R$ [England$_C$ and$_N$ Canada$_C$]$_C$]$_E$'' 
\item
``I may have forgotten my keys [on$_R$ [[the table]$_C$ or$_N$ couch$_C$]$_C$]$ _A$''
\end{itemize}

\paragraph{Vocatives.} Vocatives are generally G, as they are exclusively part of the speech event Scene.
If the person addressed is also a participant, we mark is as an A and a G.

\be 
\item
``[John$_G$, who$_A$ is$_F$ this$_A$ ?]$_H$''
\item  ``[John$_{G+A}$, go$_P$ outside$_A$]$_H$'' 
\german{\item
``[Nein$_H$, [Herr Kapitan]$_G$]$_H$''}
\ee

\paragraph{Titles.} By convention, titles of names are considered Elaborators of the proper name.

\be 
\item
``I$_A$ can$_D$ 't$_D$ find$_P$ [Captain$_E$ Nemo$_C$]$_A$''
\item
``[Queen$_E$ Mary$_C$]$_A$ went$_D$ to$_F$ sleep$_P$''
\ee

\paragraph{Dates and addresses.} We use multiple Cs to internally mark the different parts of such expressions: 

\be
\item day/month/year: 
\begin{itemize}
\item
``John will arrive$_P$ [on$_R$ November$_C$ 5$_C$ , 2019$_C$]$_T$''
\item
``John and Mary are due to meet [today$_C$ [at$_R$ 16:00$_C$]$_C$]$_T$''
\end{itemize}

 \item city/state/country:
\begin{itemize}
\item
``Springfield$_C$, Ohio$_C$, [United States]$_C$''
\end{itemize}

\item
street/house number/apartment number:
\begin {itemize}
\item 
``[31]$_C$ [Herzl St]$_C$,  [Apt$_C$ 207$_E$]$_C$, [Tel Aviv-Yafo]$_C$''
\end {itemize}
\ee

\paragraph{Question Words.} Question words should be annotated with the same category as their respective component in a given answer. 

\be \item
``How$_D$ did you fix your car?''
\item
``Who$_A$ shot the sheriff?''
\item
``[Which$_E$ car$_C$]$_A$ did you buy?''
\item
``Why$_H$ [haven't you called me]$_H$ ?''
\item
``When$_T$ will they arrive?''
\ee

The same applies to indirect questions, where the question word is elaborated by a scene in which it serves as a remote participant:

\be
\item 
``Tell$_P$ me$_A$ [what$_C$ [\rem{what}$_A$ happened$_P$]$_E$]$_A$''
\item
``I know [what$_C$ [you gave$_P$ \rem{what}$_A$ [to$_R$ John$_C$]$_A$]$_E$]$_A$''
\item
``I$_A$ wonder$_P$ [where$_C$ [he is going \rem{where}$_A$]$_E$]$_A$''

\ee

\noindent
Some of these words can also be used as relative pronouns modifying a noun. In such cases they are not real questions,
but relators of the E Scene with the elaborated entity, so they should be marked as Rs.

\be
\item
``the$_F$ man$_C$ [who$_R$ was$_F$ n't$_D$ there$_{S}$ \rem{man}$_A$]$_{E}$''

\item
``the$_F$ tiger$_C$ [which$_R$ lost$_P$ [his$_E$ hair$_C$]$_A$ \rem{tiger}$_A$]$_{E}$''
\item
``the$_F$ city$_C$ [in$_R$ which$_R$ John$_A$ lives$_S$ \rem{city}$_A$]$_{E}$''
\ee

\paragraph{Non-contiguous Linkers.} 
Sometimes a linkage relation is expressed by several words, which are not contiguous in the text, but evoke a single relation. We mark them by convention as two separate linkers and not as a non-contiguous unit.
\be \item
``[Either]$_L$ you buy it [or]$_L$ you don't''
\ee


\paragraph{Reflexives.} Reflexives are the words that (in their primary sense) state that two participants of an event are one and the same (``himself'', ``themselves'', ``to one another'' etc.). In UCCA, we mark them as part of the P/S, which is considered unanalyzable. Note, however, that in some cases reflexives are not used in their primary sense. In these cases, they should be analyzed according to their meaning in the context.
\be \item
``John$_A$ [washed himself]$_P$''
\item
``Mary$_A$ [talked herself]$_P$ [into coming]$_A$''
\item
``John$_A$ [looked at himself]$_P$ [in the mirror]$_A$''
\item
``[He$_C$ himself$_F$]$_A$ spoke$_P$ [to the manager]$_A$.'' (``himself'' here does not introduce a participant, but rather emphasizes that it was ``he'' and not someone else)
\item
``He did it [all$_E$ [by$_R$ himself$_C$]$_C$]]$_D$'' (it's a D since the expression basically means that he did it alone)
\item
``John$_A$ [relieved himself]$_P$ [in$_R$ the$_F$ backyard$_C$]$_A$''
\item
``John$_A$ [established himself]$_P$ [as$_R$ a$_F$ lecturer$_C$]$_A$''
\german{
\item
``John hat$_F$ [sich gewaschen]$_P$''}
\german{
\item
``[Studieren$_P$]$_A$ [lohnt sich]$_P$''
}
\ee

\paragraph{Complex Prepositions.} Some prepositions are multi-worded. They should be annotated as complex units (or as unanalyzable if they have no parts with significant semantic input). \textcolor{red}{In German this could be ``auf Grund'', ``an der Seite von'', ``des Weiteren'' etc.}
\be \item
``[According to]$_P$ John$_A$, [ [the$_F$ soup$_C$]$_A$ is$_F$ salty$_S$]$_A$''
\item
``Mary$_A$ is$_F$ [on top of]$_S$ [this$_E$ task$_C$]$_A$''
\item
``[[later in]$_R$ 1988$_C$]$_T$, John$_A$ bought$_P$ [a$_F$ car$_C$]$_A$''
\ee



\paragraph {Coordinated Main Relations.} 

\begin {enumerate}
\item
If two or more coordinated Processes or two or more coordinated States share exactly the same scene elements (As, Ds, Ts), then they should be united under one parent unit which will be assigned two categories: Coordinated Main Relation (CMR) and also P or S accordingly. Internally, we will use multiple Cs to mark each Main Relation separately. 

\begin{itemize}
\item 
``[John$_A$ [wrote$_C$ and$_N$ recorded$_C$]$_{CMR+P}$ [a song]$_A$ [on Tuesday]$_T$]$_H$''
\item
``John$_A$ is [quiet$_C$ and$_N$ shy$_C$]$_{CMR+S}$''
\end{itemize}

\item
On the other hand, if the coordinated Main Relations do not share one of the Scene elements, we should  separate them into different scenes.
\begin{itemize}
\item
``[John$_A$ [woke up]$_P$]$_H$, [went$_P$ [to school]$_A$ \rem{John}$_A$]$_H$ and$_L$ [met$_P$ Mary$_A$ there$_A$ 
\rem{John}$_A$]$_H$''
\end{itemize}
\end{enumerate}



\paragraph{Sequences of Multiple Parallel Scenes}

Multiple Parallel Scenes often form a continuous sequence of events (with or without Linkers). Such a string of events need not be placed under one parent Parallel Scene unit. 

\begin{itemize}
\item 
``[John went to the store]$_H$ to$_L$ [buy eggs \rem{John}]$_H$ but$_L$ [unfortunately the store was closed]$_H$'' (The three linked Hs in this case should not be united under one parent H)
\end{itemize}

\paragraph{Cross-sentence annotation}

\begin{itemize}
\item
Quotation that spans multiple sentences: \\
``[John$_A$ said$_P$ [to Jake]$_A$: [`[I noticed you weren't here yesterday]$_H$. [Would you like me to update you on recent decisions?]$_H$']$_A$]$_H$'' 

\textit{Just the structure:} [$A$ $P$ $A$: [`$H$ $H$']$_A$]$_H$

\item
Cross-sentence Remotes: \\
``[Jane asks Mary [if she has heard the evening news]$_A$]$_H$. [Mary says she hasn't \rem{heard}$_P$ \rem {news}$_A$]$_H$ but$_L$ [\rem{she}$_A$ will [make sure]$_D$ to$_F$ \rem{heard}$_P$ \rem{news}$_A$]$_H$ when$_L$ [she arrives home]$_H$''

Note that but$_L$ and when$_L$ are at the top level even though they link parallel scenes within a sentence.
That is, Hs linked together within a sentence need not be nested in a larger parallel scene.

\item
Imperatives: ``[\rem{You}$_A$ Eat]$_H$. [You$_A$ 'll feel better]$_H$.''
\item
Interjections: ``[Great$_G$ ! I'm happy it worked out]$_H$.''
\end{itemize}

\subsection{Morphology.}

\paragraph{Inflectional and Derivational Morphology.} UCCA does not annotate them in the current layer. Therefore the word ``dogs'' has no sub-unit ``dog'', and neither does the word ``talked'' have a sub-unit ``talk''. 

\paragraph{Compounds.}
Compounds written as one word should generally be split in UCCA (see Section~\ref{sec:compounds}).
However, some compounds have obtained their own idiosyncratic meaning, and it is no longer clear how the meaning of the component words contribute to the meaning of the compound. In this layer of UCCA they should be analyzed as a single unit, without sub-units.

\be \item
``There are pickpockets$_A$ in this side of town''
\item
``he$_A$ 's$_F$ [a$_F$ has-been$_C$]$_S$''
\item
``Let's$_D$ go$_P$ [to$_R$ the$_F$ merry-go-round$_C$]$_A$''
\ee

\subsection{Problematic Cases for the Scheme}

We acknowledge that comparatives, superlatives and similes are not covered well by UCCA's foundational layer categories. When you encounter one of these please place a comment to that effect in the unit containing it.

\paragraph {Superlatives:}

\be 
\item
``China$_A$ is$_F$ the$_F$ greatest$_S$''
\item
``China$_A$ is$_S$ [the$_F$ [greatest$_S$ \rem{place}$_A$]$_E$ place$_C$ [on earth]$_E$]$_A$''\\
``This$_A$ is$_S$ [[my$_S$ \rem{sharpener}$_A$]$_E$ [best$_S$ \rem{sharpener}$_A$]$_E$ sharpener$_C$]$_A$''
\item
``Mary$_A$ is$_F$ [John$_C$ 's$_R$]$_A$ best$_D$ friend$_S$''
\ee

\paragraph{Comparatives.} If the set of entities the comparison applies to is explicitly mentioned, it should be marked as a separate scene.
\be 
\item
``[James$_A$ is$_F$ taller$_S$]$_H$ than$_L$ [John$_A$ \rem{taller}$_S$]$_H$''
\item
``[James is better$_D$ [at skating]$_P$]$_H$ than$_L$ [John$_A$ \rem{skating}$_P$]$_H$''
\item
``[James runs$_P$ faster$_D$]$_H$ than$_L$ [John \rem{runs}$_P$ \rem{faster}$_D$]$_H$''
\item
``[James$_A$ is$_F$ more$_D$ interesting$_S$]$_H$ than$_L$ [John$_A$ \rem{interesting}$_S$]$_H$''
\item
``[James$_A$ is$_F$ as$_D$ competent$_S$]$_H$ as$_L$ [anyone$_A$ \rem{competent}$_S$]$_H$''
\ee

\highlight{\paragraph{Similes.} In most cases similes should be treated as separate Scenes:
\be
\item
``[He$_A$ eats$_P$]$_H$ like$_L$ [a horse \rem{eats}$_P$]$_H$''
\ee

An exception would be when the verb does not evoke a Scene of its own (e.g. ``looks'', ``seems''---see Section~\ref{sec:copula_verbless}, item~\ref{seem}) and is therefore considered a G. Then the whole phrase should be marked as one Scene:
\be
\item
``He$_A$ looks$_G$ like$_F$ [a horse]$_S$''
\item
``He$_A$ looks$_G$ like$_F$ he$_A$ just$_T$ saw$_P$ [a dinosaur]$_A$''
\ee}

\section{Criteria for compound splitting in German}\label{sec:compounds}

Some of the examples are adapted from \citet{schulte_im_walde-16}. This section is co-authored with Jakob Prange and Nathan Schneider.

\paragraph{Criterion 1:} {\bf Is the compound semantically transparent or opaque?}

\be
\item
Split transparent compounds. 
\begin{itemize}
\item
The meaning of {\it Ahornblatt} (maple leaf) can be derived from the meanings of {\it Ahorn} (maple) and {\it Blatt} (leaf).
\end{itemize}

\item
Don't split opaque compounds.
\begin{itemize}
\item
The meaning of {\it Maulwurf} (mole) 	cannot be derived from the meanings of {\it Maul} (mouth of an animal) and {\it Wurf}\, (throw).
\end{itemize}

\item
Don't split partially/asymmetrically transparent compounds.

\begin{itemize}
\item
The meaning of {\it Zeitungsente} (newspaper hoax) cannot be derived from the meaning of {\it Ente} (duck), but it can be derived from the meaning of {\it Zeitung} (newspaper).
\item
{\it Murmeltier} (marmot) is a {\it Tier} (animal) but it does not involve either the noun {\it Murmel} (marble) or the verb {\it murmeln} (murmur).
\item
{\it Sonnenk\"onig} (``Sun King'', aka King Ludwig XIV) is a {\it K\"{o}nig} (king), but it doesn't involve a {\it Sonne} (sun). It's more of a name, and hence should not be split.
\item
{\it Geduldsfaden} (thread of patience) refers to the extent of one's patience, but doesn't involve a thread. Note that this is different from the metaphorical use of {\it Faden} (thread) as part of a conversation. Also, you cannot paraphrase it with {\it Faden der Geduld}, cf. Criterion 2.
\item
{\it Schriftzug} (logo) refers to something written ({\it Schrift} = writing), but it doesn't have to be an actual hand movement {\it Zug} (stroke) anymore, although it is derived from that originally.
\end{itemize}
\ee

\paragraph{Criterion 2:} {\bf Can the compound be paraphrased as a noun phrase with the same noun head?}

If it can be paraphrased, it should be split.

\begin{enumerate}
\item
{\it Kaufleute} (salesmen) $\rightarrow$ {\it Leute, die kaufen und verkaufen} (people that buy and sell).
\item
{\it Kinderbuch} (children's book) $\rightarrow$ {\it ein Buch f\"{u}r Kinder} (a book for children)
\item
{\it spindelf\"{o}rmig} (spindle-shaped) $\rightarrow$ {\it hat die Form einer Spindel} (has the shape of a spindle)
\end{enumerate}

{\bf Note:} Even if the head of the compound is a metaphor, if the same metaphor can be used in a paraphrase, the compound is considered compositional and should be split: {\it Bergkette} $\rightarrow$ {\it eine Kette von Bergen} (a chain of mountains), even though it's not an actual chain, but rather a chain-like arrangement of mountains.

\paragraph{Criterion 3:} {\bf Is the pattern of the compound productive? That is, can one or both of the words of the compound be altered, while retaining a similar meaning?}

\begin{enumerate}
\item
If it is, it should be split.
\begin{itemize}
\item
{\it Frucht{\bf saft}}, {\it Apfel{\bf saft}}, {\it Orangen{\bf saft}} (types of juice)
\item
{\it Schiffs{\bf herr}} (ship owner), {\it Haus {\bf herr}} (house owner)
\item
{\it Braun{\bf b\"{a}r}}, {\it Schwarz{\bf b\"ar}}, {\it Grizzly{\bf b\"ar}} (different species of bears);
BUT: {\it Waschb\"{a}r} (raccoon), {\it Armeisen{\bf b\"{a}r}} (anteater) should not be split.
\item
{\it Gebirgs{\bf zug}} (mountain range), {\it Sieges{\bf zug}} (triumphal march), {\it Vogel{\bf zug}} (bird migration) are all related,
BUT: {\it Schrift{\bf zug}} (logo) doesn't have much to do with the above compounds and should not be split.
\end{itemize}
\item
Where one of the words of the compound cannot be used as a free word, or has a very different meaning when used that way, it should not be split.
\begin{itemize}
\item
{\it Uhrwerk}, {\it Fachwerk}, {\it Triebwerk}, {\it Schuhwerk}, {\it Blattwerk} are all related, BUT {\it Werk} is an opus, a piece of art or a factory and therefore should not be split (borderline).
\end{itemize}
\end{enumerate}

\setcounter{secnumdepth}{0}
\section{Appendix: Plain Text Notation}

In order to make UCCA's annotation legible and standardized, we give here guidelines for UCCA's notation in plain text. We note that the hierarchical structure formed by UCCA can be annotated by standard bracketing. The abbreviation of the category should be either adjacent to the left or to the right side of the category.
For example, annotating the word ``apple'' with the category X should look like ``[X apple]'' or ``[apple X]''.
We use the following abbreviations for the categories:\\

\noindent
T -- time \\
Q -- quantifier\\
H -- parallel Scene \\
A -- participant \\
C -- center \\
L -- linker\\
D -- adverbial\\
E -- elaborator\\
G -- ground\\
S -- state\\
N -- connector\\
P -- process\\
R -- relator\\
F -- function\\


\paragraph{Non-contiguity:} We use a dash to indicate a continuation of a unit. For example, if ``word1 $\ldots$ word2'' is a non-contiguous unit then we mark it ``[X- word1] [Y] [Z] [W] [-X word2]''.

``[John A] [P- took] [Mary A] [up on -P] [ [her A] [promise P ] A]''

In case there are two non-contiguous units nested within one another, and of the same category, we may use indices to disambiguate. For example, in the sequence ``w1 w2 w3 w4 w5'', if ``w1 \ldots w4'' is a non-contiguous unit of category X and ``w2 \ldots w5'' is also a non-contiguous unit of category X, we mark it ``[X1- w1] [X2- w2] w3 [-X1 w4] [-X2 w5]''.

\paragraph{Remote Units:}
We place the Remote unit inside its Parent unit at the end of the phrase in round brackets and assign it with the relevant category:
\begin{itemize}
\item
``John got home and [took a shower \rem{John}$_A$]''
\end{itemize}

\paragraph{Implicit Units:}
Implicit units are marked much like remote units, the only difference is that we add a fixed expression ``IMP'' inside the round brackets. 
\begin{itemize}
\item
``[Not going there any more \imp$_A$]'' (Who is not going there is implicit)
\end{itemize}

\section{Appendix: Restrictions Summary}

\begin {enumerate}
\item Kinds of units:
\begin {enumerate}
\item Superparallel unit (main element: H). This includes the root unit---only H and L children are allowed at the top level.
\item Scene unit (main element: P or S)
\item Sub-scene unit (main element: C)
    \begin{enumerate}
    \item Connective (C, N; may not have E, Q)
    \item Non-connective (C; may have E, Q)
    \end{enumerate}
\item Lexical unit (tokens only)
\end{enumerate}

\item Scene categories:

\begin {enumerate}
\item
Each scene unit has to include either a Process (P) or State (S) as the main relation.

Other optional scene elements:

\item Participant (A)
\item Adverbial (D)\footnote{As an exception see Section~\ref{D}, ``\nameref{D}'': although D is a scene category, in this particular case it can be marked inside a non-scene unit.}
\item Time (T)
\item Ground (G)
\item Coordinated Main Relation (CMR): can be added only as a secondary category to either a P or an S. 
\end{enumerate}

\item Non-scene categories: 

\begin{enumerate}
\item 
Center (C):  unless a non-scene unit contains only one token or is unanalyzable it will need to include at least two child units one of which should be C.

Other optional non-scene units:
\item Elaborator (E)
\item Connector (N)
\item Quantifier (Q)
\end{enumerate} 

\item
Categories that can appear both in scene and non-scene units: 
\begin{enumerate}
\item Relator (R)
\item Function (F)
\end{enumerate}

\item Other categories: 

\begin {enumerate}
\item Parallel Scene (H) 
\item Linker (L) - links only between Hs

Note that all top layer units are either Hs or Ls

\item Unanalyzable (UNA):  apart from being assigned a main category from the list above, any unit can also be assigned UNA as a secondary category. 

\end{enumerate} 

\item Remotes: 

\begin{enumerate}
\item
F should not be added as remote, nor should an F unit have remote children.

\item
For a unit to have children, at least one of them must be non-remote and must have a category other than F.
\end{enumerate}

\item Types of scene units (units that can be internally annotated as scenes):
\begin {enumerate}
\item Participant (A) scene 
\item Elaborator (E) scene 
\item Center (C) scene (should be used sparingly)
\item Parallel Scene (H): if it is not an A-scene, E-scene or C-scene it is necessarily an H. Note that a sequence of Hs (with or without Ls) should not be united under another H.
\end{enumerate}

\end{enumerate}

\section{Appendix: Interesting Examples}
\dcom{\nss{Additional constructions: specific genres like diary style (``Went to the beach.''---implicit `I' subject) and labelese (``Contains nuts.''---implicit subject; ``Use with caution.''---imperative plus implicit object). see \citet{ruppenhofer-10}}\oa{we can add the latter two to the interesting examples chapter}

\nss{What about `a lot (of food)', `a little (confused)', etc.?}\oa{analyzable imo, since ``a little'' has to do with ``little''}}

\paragraph {Quantity modifiers, distances, and directions}

\begin{itemize}
\item
``John$_A$ lives$_S$ [[3$_Q$ miles$_C$]$_Q$ away$_R$ from$_R$ DC$_C$]$_A$''
\item
``John$_A$ moved$_P$ [to$_R$ [about$_E$ 3$_Q$ miles$_C$]$_Q$ away$_R$ \imp$_C$]$_A$''
\item
``The cat appeared$_P$ [from$_R$ behind$_R$ the$_F$ couch$_C$]$_A$''
\end{itemize}

\paragraph {Instruments}
\begin{itemize}
\item
``He$_A$ broke$_P$ [the window]$_A$ [with a hammer]$_A$''
\item
``He$_A$ broke$_P$ the window [using$_R$ a$_F$ hammer]$_A$''
\item
``He broke the window [by$_R$ using$_R$ a$_F$ hammer$_C$]$_A$''
\item
``[He$_A$ used$_P$ [a$_F$ hammer]$_A$]$_H$ to$_L$ [break$_P$ [the$_F$ window$_C$]$_A$ \rem{he}$_A$]$_H$''
\end{itemize}

\paragraph{Postmodifiers}

\begin{itemize}
\item 
``He$_A$ arrived$_P$ [five$_Q$ minutes$_C$ \textbf{ago$_R$}]$_T$''
\item
``He$_A$ is$_F$ [nine$_Q$ years$_C$]$_T$ \textbf{old$_S$}''
\item
``Does$_F$ [everyone$_C$ \textbf{else$_E$}]$_A$ want$_P$ pizza$_A$?''
\end{itemize}

\paragraph{Nested Fused E-Scenes.}
\begin{itemize}

\item
``[how$_C$ [[what$_C$ [we$_A$ 've$_F$ been$_F$ building$_P$ \rem{what}$_A$]$_E$]$_A$ is$_F$ [of use]$_S$ [for$_R$ the$_F$ field$_C$]$_A$ \rem{how}$_D$]$_E$''
\end{itemize}

\paragraph{Quantities without Thing Quantified.}
Such cases requires adding remote units.

\begin{itemize}
\item
``[[most$_Q$ \rem{books}$_C$]$_C$ [of$_R$ the$_F$ books$_C$]$_E$]$_A$ are missing'' (Contrast: [most$_Q$ books$_C$]$_A$ are missing)
\item
``[[4$_Q$ \rem{books}$_C$]$_C$ [of$_R$ the$_F$ 5$_Q$ books$_C$]$_E$]$_A$ are missing''
\item
``[[1$_Q$ \rem{books}$_C$]$_C$ [of$_R$ the$_F$ 5$_Q$ books$_C$]$_E$]$_A$ is missing'' (note the singular/plural mismatch between the primary and remote descriptor)
\end{itemize}


\paragraph{Degree-consequence constructions}

\begin{itemize}
\item 
``[John was [so$_E$ good$_C$]$_D$ [in$_R$ faking$_C$]$_P$ [his mother's signature]$_A$]$_H$ that$_L$ [the teacher couldn't tell the difference]$_H$''
\item
``[John was so$_D$ tired$_S$]$_H$ that$_L$ [he forgot to pick up his brother from school]$_H$''
\item
``[She was [in$_R$]$_{P-}$ such$_D$ [a hurry]$_{-P}$]$_H$ that$_L$ [she forgot to take her keys]$_H$''
\item
``[John is sensible$_S$ enough$_D$]$_H$ to$_L$ [know the difference between right and wrong \rem{John}$_A$]$_H$''
\item
``[Mary was too$_D$ tired$_S$]$_H$ to$_L$ [call John \rem{Mary}$_A$]$_H$''
\end{itemize}

\paragraph{Coordination among scene and non-scene units}

\begin{itemize}
\item 
``[he only cares about money]$_H$ and$_L$ [[going home on time]$_A$ (he)$_A$ (only)$_D$ (cares)$_P$]$_H$''
\end{itemize}

\dcom{\section{Appendix: Possible Post-processing Notes}

\nss{examples would help! I've attempted a couple}

\begin{itemize}
\item
  In E Scenes, put the Cs elaborated on in the Elaborator Scene.
\item
  In Ground, extract the G from the Scene they are positioned in, and add a root node whose children are the G and
  the Scene.
\item
 Flag: turn all the Rs into Fs, especially if a PSS layer is included.
\item
  Include determiners within the main relation if they are in an A-Scene noun phrase.\\ 
 \hspace*{1em} I remember$_P$ [the$_F$ party$_P$]$_A$ $\Rightarrow$ I remember$_P$ the$_F$ [party$_P$]$_A$ \\
 \hspace*{1em} I remember$_P$ [\rem{I}$_A$ starting$_D$ a$_F$ meal$_P$]$_A$ $\Rightarrow$ I remember$_P$ [\rem{I}$_A$ starting$_D$]$_{A-}$ a$_F$ [meal$_P$]$_{-A}$
\item
  Possessive pronouns should be S+A
\item
  Negative polarity relators (without, neither) should be annotated both as negation \nss{D?} and as L/R. \nss{although in another place it says R is residual, so other categories take priority}
 \item 
  Articles in English should always be F
  \item
  Instead of marking it D, German ``Kein'' and its declensions should be Q (with the modified noun being the C), the parent A should be marked for negation: ``ich habe [keinen$_Q$ Stift$_C$]$_{A+D}$''
\end{itemize}}

\bibliographystyle{plainnat}
\bibliography{uccaguidelines}

\end{document}